\documentclass{article}

\usepackage{microtype}
\usepackage{graphicx}
\usepackage{subfigure}
\usepackage{booktabs} 
\usepackage{caption}
\usepackage{hyperref}

\usepackage[dvipsnames]{xcolor}
\usepackage{listings}
\usepackage{float}
\usepackage{marginnote}
\usepackage{tabularx}
\usepackage{ragged2e}

\definecolor{codegreen}{rgb}{0,0.6,0}
\definecolor{codegray}{rgb}{0.5,0.5,0.5}
\definecolor{codepurple}{rgb}{0.58,0,0.82}
\definecolor{backcolour}{rgb}{0.95,0.95,0.92}

\lstdefinestyle{mystyle}{
  backgroundcolor=\color{backcolour},
  commentstyle=\color{codegreen},
  keywordstyle=\color{magenta},
  numberstyle=\tiny\color{codegray},
  stringstyle=\color{codepurple},
  basicstyle=\ttfamily\small,
  breakatwhitespace=false,
  breaklines=true,
  captionpos=b,
  keepspaces=true,
  numbers=left,
  numbersep=5pt,
  showspaces=false,
  showstringspaces=false,
  showtabs=false,
  tabsize=2 }
\newfloat{lstfloat}{tbp}{lop}
\floatname{lstfloat}{Listing}

\usepackage[accepted]{mlsys2022}

\usepackage{amsmath,amsthm,amsfonts,bm}
\usepackage{tabularx,threeparttable,array,colortbl}
\usepackage{xspace}
\usepackage{calc}
\usepackage[inline]{enumitem}
\usepackage[title]{appendix}
\usepackage{dblfloatfix}

\newcommand{\defn}[1]       {{\textit{\textbf{#1}}}}
\newcommand{\ang}[1]            {\ifmmode{\mathopen{}\left\langle #1 \right\rangle\mathclose{}}
                                 \else{$\mathopen{}\left\langle${#1}$\right\rangle\mathclose{}$}\fi}

\newcommand{\punt}[1]{}

\newcommand{\secref}[1]         {Section~\ref{sec:#1}}
\newcommand{\secreftwo}[2]      {Sections \ref{sec:#1} and~\ref{sec:#2}}

\newcommand{\figlabel}[1]   {\label{fig:#1}}
\newcommand{\tablabel}[1]   {\label{tab:#1}}
\newcommand{\seclabel}[1]   {\label{sec:#1}}

\newcommand{\figref}[1]         {Figure~\ref{fig:#1}}
\newcommand{\lstref}[1]         {Listing~\ref{lst:#1}}
\newcommand{\tabref}[1]         {Table~\ref{tab:#1}}
\newcommand{\tabreftwo}[2]      {Tables~\ref{tab:#1} and~\ref{tab:#2}}

\newcounter{ccount}
\newenvironment{closeenum}
    {\begin{list}{\arabic{ccount}.}
    {\usecounter{ccount}\setlength{\itemsep}{-0.3\baselineskip}
     \setlength{\topsep}{0.15\baselineskip}
     \setlength{\parskip}{0pt}}}
    {\end{list}}

\newcommand*\makeAlph[1]{%
	  \ifnum#1<1\else
	      \ifnum#1>26 a\makeAlph{\numexpr#1-26}
	          \else\symbol{\numexpr96+#1}\fi\fi}

\newcounter{ccountletters}
\newenvironment{closeenumletters}
    {\begin{list}{\makeAlph{\arabic{ccountletters}})}
    {\usecounter{ccountletters}\setlength{\itemsep}{-0.3\baselineskip}
     \setlength{\topsep}{0.15\baselineskip}
     \setlength{\parskip}{0pt}}}
    {\end{list}}

\MakeRobust{\Call}
\algnewcommand{\LineComment}[1]{\State \(\triangleright\) #1}
\newcolumntype{Y}{>{\centering\arraybackslash}X}

\graphicspath{{figs/}}

\newcommand{\origsystem}{PyG\xspace}
\newcommand{\paperdata}[1]{#1}
\newcommand{\oursystem}{SALIENT\xspace}
\newcommand{\neighborsampler}{\texttt{NeighborSampler}\xspace}
\newcommand{\real}{\mathbb{R}}

\newcommand{\clearpagedraft}{}

\definecolor{Gray}{gray}{0.9}

\algnewcommand\algorithmicforeach{\textbf{for each}}
\algdef{S}[FOR]{ForEach}[1]{\algorithmicforeach\ #1\ \algorithmicdo}

\makeatletter
\newcommand{\multlinealgo}[1]{%
  \begin{tabularx}{\dimexpr\linewidth-\ALG@thistlm}[t]{@{}X@{}}
    #1
  \end{tabularx}
}
\makeatother

\newcolumntype{L}{>{\raggedright\arraybackslash}X}
\newcolumntype{R}{>{\raggedleft\arraybackslash}X}
\newcolumntype{C}{>{\centering\arraybackslash}X}

\mlsystitlerunning{Accelerating Training and Inference of Graph Neural Networks with Fast Sampling and Pipelining}

\begin{document}

\twocolumn[
\mlsystitle{Accelerating Training and Inference of Graph Neural Networks with Fast Sampling and Pipelining}



\mlsyssetsymbol{equal}{*}

\begin{mlsysauthorlist}
\mlsysauthor{Tim Kaler}{equal,mit,mitibm}
\mlsysauthor{Nickolas Stathas}{equal,mit,mitibm}
\mlsysauthor{Anne Ouyang}{equal,mit,mitibm,ibm}
\mlsysauthor{Alexandros-Stavros Iliopoulos}{mit,mitibm}
\mlsysauthor{Tao B. Schardl}{mit,mitibm}
\mlsysauthor{Charles E. Leiserson}{mit,mitibm}
\mlsysauthor{Jie Chen}{mitibm,ibm}
\end{mlsysauthorlist}

\mlsysaffiliation{mit}{MIT CSAIL}
\mlsysaffiliation{mitibm}{MIT-IBM Watson AI Lab}
\mlsysaffiliation{ibm}{IBM Research}

\mlsyscorrespondingauthor{Jie Chen}{chenjie@us.ibm.com}

%
%
%
%
%
%

\mlsyskeywords{Graph neural network, distributed multi-GPU training, fast sampling, pipelining}

\vskip 0.3in

\begin{abstract}
Improving the training and inference performance of graph neural
networks (GNNs) is faced with a challenge uncommon in general neural
networks: creating mini-batches requires a lot of computation and data
movement due to the exponential growth of multi-hop graph
neighborhoods along network layers. Such a unique challenge gives rise
to a diverse set of system design choices.  We argue in favor of
performing mini-batch training with neighborhood sampling in a
distributed multi-GPU environment, under which we identify major
performance bottlenecks hitherto under-explored by developers: mini-batch
preparation and transfer. We present a sequence of improvements to
mitigate these bottlenecks, including a performance-engineered
neighborhood sampler, a shared-memory parallelization strategy, and
the pipelining of batch transfer with GPU computation. We also conduct
an empirical analysis that supports the use of sampling for inference,
showing that test accuracies are not materially compromised. Such an
observation unifies training and inference, simplifying model
implementation. We report comprehensive experimental results with
several benchmark data sets and GNN architectures, including a
demonstration that, for the ogbn-papers100M data set, our system
\oursystem achieves a speedup of \paperdata{3$\times$} over
a standard PyTorch-Geometric implementation with a single GPU and a
further \paperdata{8$\times$} parallel speedup with 16
GPUs. Therein, training a 3-layer GraphSAGE model with sampling fanout
\paperdata{(15, 10, 5)} takes \paperdata{2.0} seconds per epoch and
inference with fanout \paperdata{(20, 20, 20)} takes \paperdata{2.4}
seconds, attaining test accuracy \paperdata{64.58\%}.
\end{abstract}
]



\printAffiliationsAndNotice{\mlsysEqualContribution} 

\section{Introduction}
\seclabel{introduction}

Graph neural networks~(GNNs) have emerged as an important class of
methods for leveraging graph structures in machine
learning~\citep{Li2016, Kipf2017, Hamilton2017, Velickovic2018,
Xu2019, Morris2019}.  The graph structure encodes dependencies in data
representations across layers of the GNN, injecting an effective
relational inductive bias into the neural network design.  GNNs have
been shown to be successful in (semi-)supervised, unsupervised,
self-supervised, and reinforcement learning~\citep{Kipf2017, Hu2020a,
Ma2021, Mirhoseini2021} and have been applied in a number of domains
including commerce, finance, traffic, energy, and
pharmacology~\citep{Gilmer2017, Li2018, Ying2018, Weber2019,
Shang2021}.  As graph sizes continue to grow rapidly, there is a
pressing need for efficient training and inference to facilitate
further study and deployment of GNNs.

One unique challenge to GNNs
is the exponential increase of neighborhood size with respect to the
number of network layers (i.e., hops along the input 
graph)~\citep{Chen2018}.  In a typical neural-network training scenario
with stochastic gradient descent methods, e.g., Adam~\citep{Kingma2015},
computations are organized around mini-batches: a mini-batch of training
data is fed to the network to calculate the corresponding loss and
gradient, which is then used to update the model parameters.  Similarly,
for inference, input data are processed in successive mini-batches.
In GNNs, where the representation of a data point (i.e., graph node)
depends on those of its neighbors, processing a mini-batch may lead to a
prohibitively large expanded neighborhood.  Apart from the computational
cost this incurs, the features and intermediate representations of
nodes in the expanded neighborhood also consume substantial memory.
%
%
When using accelerators such as GPUs, the neighborhood data size can 
in fact exceed the accelerator memory capacity.  To mitigate this issue,
neighborhood sampling is a popular remedy, sometimes even a necessary
rescue~\citep{Hamilton2017, Chen2018, Ying2018, Zou2019, Zeng2020,
Ramezani2020, Dong2021}.

In this work, we focus on GNNs with neighborhood sampling on GPUs and 
identify batch preparation and transfer as major bottlenecks in
commonly used GNN frameworks, e.g.,
PyTorch-Geometric~(PyG)~\citep{Fey2019} and the Deep Graph
Library~(DGL)~\citep{Wang2019}.  Batch preparation entails expanding
the sampled neighborhood for a mini-batch of nodes and slicing out the
feature vectors of all involved nodes.  The corresponding subgraph and
feature vectors must then be transferred to the GPUs, since the entire graph
and feature data are often too large to fit in GPU memory.  Somewhat
surprisingly, batch preparation and transfer take substantially longer than
the core GNN training operations (loss, gradient, and model parameter
computation).  The latter are computed in the GPU and benefit from highly
optimized libraries (e.g., BLAS~\citep{Lawson1979} and
autograd~\citep{Paszke2017}).  To fully reap their benefits and
maintain high GPU utilization, the throughput of batch preparation and
transfer needs to be increased substantially.  Scaling up to use multiple
GPUs makes the need for improvement even greater.

To resolve this challenge, this work presents three performance
optimizations which are broadly applicable to current GNN architectures
and frameworks.  The first is a \emph{fast neighborhood sampler}.  We
show a principled approach to exploring the space of applicable
optimizations and identify settings which perform well across CPU
architectures.  The second is \emph{shared-memory parallelization} for
batch preparation to circumvent CPU utilization and memory bandwidth
bottlenecks present in PyG and DGL\@.  The third is \emph{pipelined
batch transfer and computation} to increase GPU utilization.

\begin{figure*}
\begin{minipage}[c]{.72\linewidth}
\centering
\subfigure[Standard PyTorch workflow.\figlabel{standard-workflow-illustration}]{
\includegraphics[width=\linewidth]{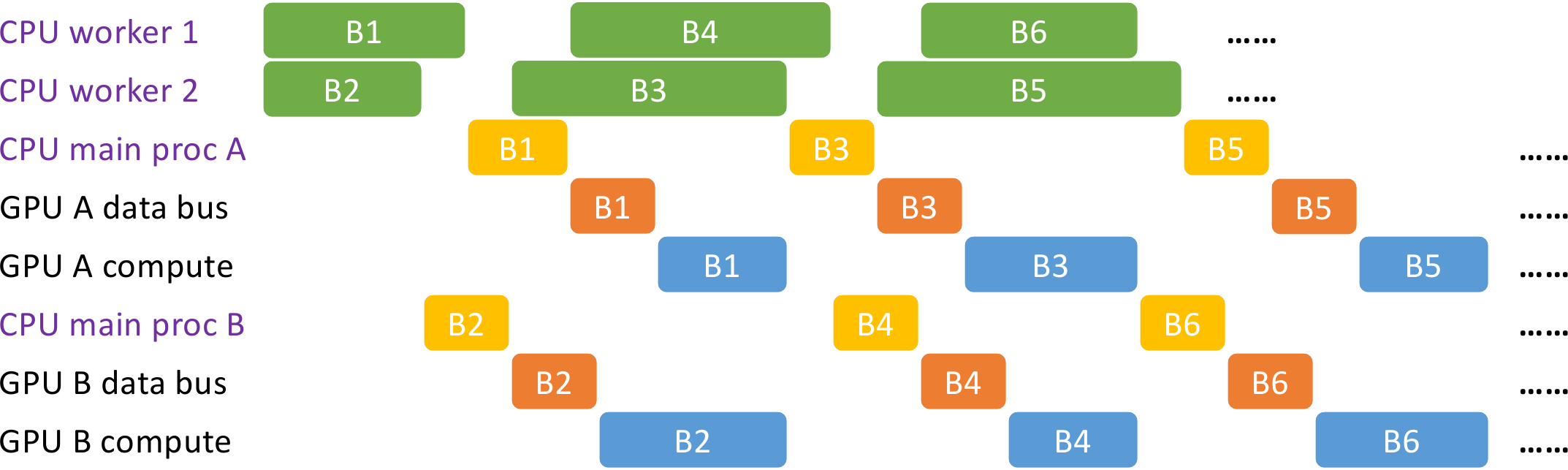}}
\subfigure[Our system \oursystem.\figlabel{salient-workflow-illustration}]{
\includegraphics[width=\linewidth]{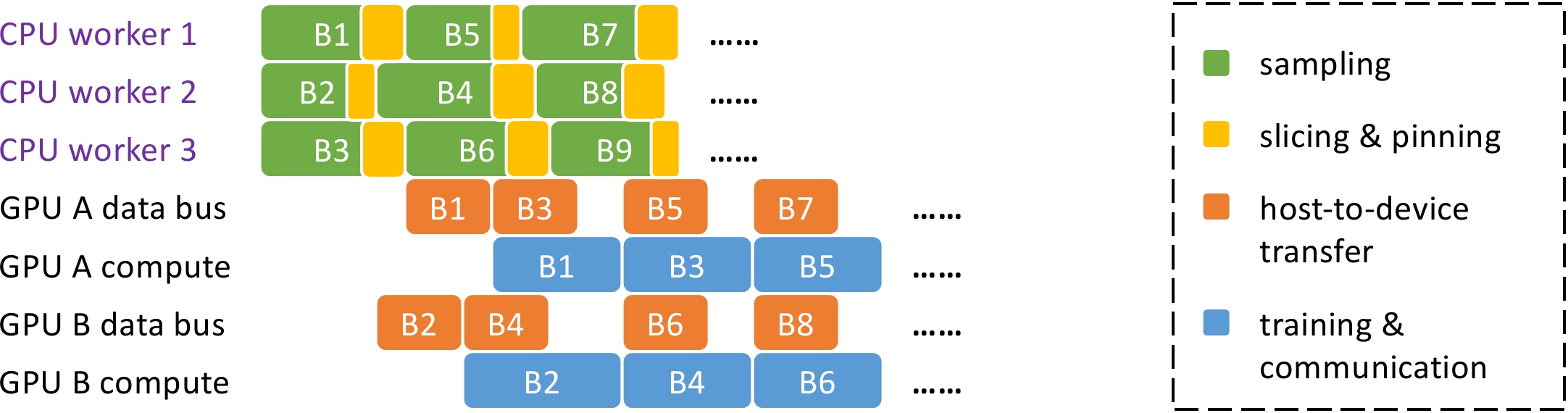}}
\end{minipage}%
\hspace*{\fill}
\begin{minipage}[c]{.25\linewidth}
\vspace*{-2em}
\caption{Illustration of mini-batch progress per training epoch: comparison
between a standard PyTorch workflow and \oursystem, the optimized
system detailed in this paper.  The $x$-axis represents elapsed
time. The ``B$i$'' blocks refer to operations with the $i$-th
mini-batch; different operations with the same batch are distinguished
by color.  In modern computing clusters, the number of available CPU
cores is often much greater than the number of GPUs, hence CPU workers
may prepare batches in parallel to try and saturate the GPUs.  With
respect to \lstref{pytorch-code}, green boxes correspond to lines
1--2, yellow boxes to lines 3--4, orange boxes to lines 5, and blue
boxes to lines 6--8.}
\label{fig:overview}
\end{minipage}
\end{figure*}

The effect of these optimizations is shown in \figref{overview},
which illustrates the timeline of mini-batch computations across CPU and 
GPU resources for a standard PyTorch workflow with and without the
optimizations.  The three optimizations respectively improve the CPU
throughput of neighborhood sampling and expansion (green boxes in
\figref{overview}); reduce slicing overhead
%
%
(yellow boxes); and enable overlapped GPU transfers and computations
(red and blue boxes).  With a reasonably high CPU-to-GPU ratio, as is
often the case in modern computing clusters, these optimizations
almost eliminate GPU idle time, enabling fast training at a
speed commensurate with that of the core training operations.

Additionally, this work studies inference.  Although trade-offs among
accuracy, speed, and memory requirements have been studied extensively
for training, they are relatively under-studied for inference.  We
conduct an empirical analysis that indicates that neighborhood
sampling in inference sacrifices prediction accuracy only
marginally.  This suggests that mini-batch inference with neighborhood
sampling is a viable alternative to layer-wise inference with full
neighborhoods, yielding accuracy comparable to the latter but with a much
lower memory footprint.  As an added advantage, model architecture code can be reused between training and inference, simplifying development.

Our system, \oursystem, addresses and alleviates bottlenecks in
\underline{SA}mpling, s\underline{LI}cing, and data movem\underline{ENT}.
\oursystem's optimizations are all done over standard GNN code written in
PyG, retaining the neural network module, the optimizer, and the
distributed data-parallel~(DDP) framework for training on multiple
machines.  This implementation minimizes adoption barriers for
developers, who can persist with their familiar deep learning frameworks and
focus on modeling and applications (e.g., experimenting with neural
architectures), without being distracted by new modules and APIs.

We highlight the following contributions:
\begin{enumerate}[leftmargin=*,topsep=0pt]
\setlength\itemsep{0em}
%
%
\item A careful analysis of GNN training codes operating on large graphs,
identifying performance bottlenecks unique to GNNs in batch
preparation and data transfer.
%
%
\item Design of an efficient batch preparation system called \oursystem that
alleviates GNN training bottlenecks with broadly applicable
optimizations to neighborhood sampling and GPU training workflows. We
show that these improvements lead to near-optimal GPU utilization.
%
%
\item An implementation of \oursystem, whose compatibility with standard PyG
code facilitates use by GNN researchers, developers, and practitioners.
%
%
\item An empirical study that suggests neighborhood sampling in
inference need not sacrifice accuracy, while reducing memory usage and
simplifying code development.
%
%
\item An evaluation of the end-to-end training performance of \oursystem on
three benchmark data sets and four GNN architectures in both single-
and multi-GPU settings. For the largest data set, ogbn-papers100M,
with a 3-layer GraphSAGE model and sampling fanout \paperdata{(15, 10,
5)}, we show a training speedup of \paperdata{3$\times$} over a
standard PyG implementation run on one GPU and a
further \paperdata{8$\times$} speedup on 16 GPUs.  Therein, training
takes \paperdata{2.0} seconds per epoch and inference with sampling
fanout \paperdata{(20, 20, 20)} takes \paperdata{2.4} seconds,
attaining test accuracy \paperdata{64.58\%}.
\end{enumerate}

 \clearpagedraft
\section{Background}
\seclabel{background}


\subsection{Graph neural networks}

The class of \emph{message passing neural networks (MPNNs)}
\citep{Gilmer2017} encompasses a large number of GNNs.  Let $G=(V,E)$ be
a graph with node set $V$ and edge set $E$.
%
%
Let $X\in\real^{n\times f}$ be the feature matrix whose rows are node
feature vectors (denoted by $x_v\in\real^f$ for node $v$). Let $\ell =
1, \ldots, L$ denote the layer index and $\mathcal{N}(v) = \{u \mid
(u,v) \in E\}$ denote the one-hop neighborhood of~$v$. MPNNs are based on
the following update rule:
\begin{equation}
\label{eqn:MPNN}
h_v^{\ell} = 
\textsc{upd}^{\ell}\Big( h_v^{\ell-1}, 
\textsc{agg}^{\ell}\big( \{ h_u^{\ell-1} \mid u \in \mathcal{N}(v) \} \big) \Big) ,
%
%
\end{equation}
where $h_v^{\ell}$ is the layer-$\ell$ representation of $v$,
$\textsc{agg}^{\ell}$ is a set aggregation function, and
$\textsc{upd}^{\ell}$ is the update function. Initially,
$h_v^0=x_v$. After $L$ layers of updates, $h_v^{L}$ becomes the final
representation.  Overall, $v$'s layer-$\ell$ representation
$h_v^{\ell}$ depends on the previous-layer representations of $v$ and its
neighbors.

GNNs differ in their design of the two functions
in~\eqref{eqn:MPNN}. For example, in GraphSAGE~\citep{Hamilton2017},
$\textsc{agg}^{\ell}$ is a mean, LSTM, or pooling operator, and
$\textsc{upd}^{\ell}$ concatenates the two arguments and applies a
linear layer.  In GIN~\citep{Xu2019}, $\textsc{agg}^{\ell}$ is the sum
of $\{h_u^{\ell-1}\}$ and $\textsc{upd}^{\ell}$ is the sum of its
arguments followed by an MLP\@.  In GAT~\citep{Velickovic2018},
$\textsc{agg}^{\ell}$ is the identity and $\textsc{upd}^{\ell}$
computes $h_v^{\ell}$ as a weighted combination of
$W^{\ell-1}h_u^{\ell-1}$ for all $u\in\{v\}\cup\mathcal{N}_v$, where
the weights are attention coefficients and $W^{\ell-1}$ is the
parameter matrix of the layer.
\subsection{Neighborhood sampling}
From~\eqref{eqn:MPNN}, one sees that computing $v$'s representation
requires recursively inquiring neighbors, which may incur a
prohibitively large $L$-hop neighborhood; similarly for a mini-batch
of nodes.  Restricting the neighborhood size via sampling proves 
to be an effective training strategy for improving memory and time
efficiency.  Current sampling approaches generally fall under three
categories: node-wise sampling, layer-wise sampling, and subgraph
sampling.

Node-wise sampling approaches, including
GraphSAGE~\citep{Hamilton2017} and PinSage~\citep{Ying2018}, modify
the neighborhood $\mathcal{N}(v)$ in~\eqref{eqn:MPNN} by taking a
random subset containing at most $d$ neighbors, sampled without
replacement.  It is typical to specify a different sample size (called
\emph{fanout}), $d^{\ell}$, for each layer $\ell$.  The fanout
parameters serve as an upper bound on the effective degree during
neighborhood expansion.
%

Layer-wise sampling approaches collect the neighbors of all nodes in a
mini-batch and then sample the entire neighborhood for the batch.
Sampling proceeds recursively layer by layer.  Representative
methods are FastGCN~\citep{Chen2018} and LADIES~\citep{Zou2019}. These
approaches impose a nontrivial sampling distribution over the
neighborhood and rescale the neighbor representations through dividing them
by their respective sampling probability, to preserve unbiasedness
before activation.  Nonlinear activation functions will destroy
unbiasedness anyway, but training convergence results can still be
established based on asymptotic consistency~\citep{Chen2018a}.

Subgraph sampling approaches, such as Cluster-GCN~\citep{Chiang2019}
and GraphSAINT~\citep{Zeng2020}, sample a connected subgraph and
compute mini-batch loss restricted to this subgraph.

There exist other types of sampling-related approaches as
well. Authors of LazyGCN~\citep{Ramezani2020} study the promise of
lowering sampling frequency and propose a ``lazy'' sampling schedule
that is applicable to all of the above categories.  Inspired by
LazyGCN, authors of GNS~\citep{Dong2021} further propose caching a
random but sufficiently large subgraph, from which node-wise sampling
is performed for each training epoch.

\subsection{GNN training systems}

At the system level, due to the unique characteristics of GNNs, many
efforts have been made to develop training systems scalable to large
graphs, based on either mainstream deep learning frameworks or more
specialized systems.

Roc~\citep{Jia2020} and DeepGalois~\citep{Hoang2021} are examples of
the latter, both of which perform full-batch training as opposed to
mini-batch. Also perform full-batch training are
  NeuGraph~\citep{Ma2019}, which is based on
  TensorFlow~\citep{Abadi2015}; and FlexGraph~\citep{Wang2021},
  Seastar~\citep{Wu2021}, and GNNAdvisor~\citep{Wang2021a}, which are
  based on PyTorch~\citep{Paszke2019}. On the other hand,
  DistDGL~\citep{Zheng2020}, Zero-Copy~\citep{Min2021},
  GNS~\citep{Dong2021}, and $P^3$~\citep{Gandhi2021} are based on
  PyTorch and the DGL module and all perform mini-batch training. In
the referenced publications, the authors report results on multiple
machines (CPUs only), single machines with multiple GPUs, or single
machines with a single GPU.  Our system, \oursystem, is based on
PyTorch and the PyG module and performs mini-batch training. We
demonstrate results on a single machine with a single GPU as well as
multiple machines with multiple GPUs each.

   \clearpagedraft
\section{Performance Characteristics of Neighborhood Sampling in GNNs}
\seclabel{overview}

This section summarizes our investigation into the performance
bottlenecks in standard implementations of GNNs in PyTorch.  Our
findings underscore the gap between hardware capabilities and
actualized performance and motivate the optimizations in \oursystem, 
which are detailed in \secref{salient}.

For this performance study, we use as reference a standard $3$-layer 
GraphSAGE architecture implemented in PyG, running on a $20$-core Intel 
Xeon Gold 6248 CPU and a single NVIDIA Volta V100 GPU.  At a high 
level, the baseline implementation for our study includes the following
operations, with corresponding pseudocode in \lstref{pytorch-code}:

\begin{closeenum}
\item \textbf{Batch preparation:} Sample a multi-hop neighborhood for
  a given mini-batch, and slice features and label tensors to obtain
  subtensors for nodes in the sampled neighborhood. (Lines 1--4)
\item \textbf{Data transfer:} Transfer the prepared batch (a sampled
  neighborhood and sliced tensors) to the GPU\@. (Line 5)
\item \textbf{GPU training:} Perform model evaluation, back
  propagation, and model update on the GPU\@. (Lines 6--8)
\end{closeenum}

The baseline \origsystem code was written to be a good representation of a 
performance-tuned code using standard libraries.  It includes the
following conventional optimizations:
\begin{enumerate*}[label=(\roman*)]
\item row-major representation of the feature matrix
to improve CPU cache efficiency in slicing operations;
\item CPU-to-GPU transfers via pinned memory to enable asynchronous
transfer with direct memory access; and
\item half-precision floating point for feature vectors in host memory to
reduce bandwidth pressure in slicing and CPU-to-GPU data transfers (GPU 
training computations are still done in single precision).
\end{enumerate*}
In our experiments, these optimizations yield a roughly 2$\times$ speedup per
epoch over a naive PyG implementation of \lstref{pytorch-code} or about 
1.5$\times$ over a reference DGL benchmark.%
\footnote{GitHub repo: \href{https://github.com/dglai/dgl-0.5-benchmark}{dglai/dgl-0.5-benchmark}.}
The resulting code, hereafter referred to simply as ``PyG,'' serves as the
baseline for our performance evaluations and the starting point for
\oursystem.


\begin{lstfloat}
\begin{lstlisting}[language=Python]
ns = NeighborSampler(G, fanouts, batch_sz)
for Gs, ids in ns: # A sampled subgraph Gs
  xs, ys = x[ids],y[ids[:batch_sz]] # Slice
  batch = (xs, ys, Gs)
  batch = batch.to(GPU) # Transfer to GPU
  optimizer.zero_grad() # Train on GPU
  loss_fn(model(batch), ys).backward()
  optimizer.step()
\end{lstlisting}
\vskip -0.1in
\caption{Reference pseudocode for a standard PyTorch implementation of 
  GNN training with neighbor sampling on graph $G$ with node features $x$ 
  and labels $y$.}
\label{lst:pytorch-code}
\end{lstfloat}

\subsection{Observed per-operation performance}

\begin{table}
  \centering
  %
  %
  \caption{Per-operation performance breakdown of the baseline PyG training 
    code.  Reported runtimes correspond to blocking or non-overlapped 
    computations among the steps outlined in \lstref{pytorch-code}.  GNN:
    $3$-layer GraphSAGE with fanouts (15,10,5), hidden-layer feature
    dimensionality 256, and mini-batch size 1024.  Data sets are introduced 
    in \secref{eval}.}
  \tablabel{bottleneck}
  \vskip 0.05in
  {\small
  \tabcolsep 5.0pt
  \begin{tabularx}{\linewidth}{@{} Xrrrrrrr @{}}
    \toprule
    \emph{Data Set}
    & \multicolumn{1}{c}{\emph{Epoch}}
    & \multicolumn{2}{c}{\emph{Batch Prep.}}
    & \multicolumn{2}{c}{\emph{Transfer}}
    & \multicolumn{2}{c@{}}{\emph{Train (GPU)}}
    \\
    \cmidrule(lr){2-2}\cmidrule(lr){3-4}\cmidrule(lr){5-6}\cmidrule(l){7-8}
    & \emph{time} & \emph{time} & \% & \emph{time} & \% & \emph{time} & \%
    \\
    \midrule
    arxiv     & 1.7s    & 1.0s  & 58\%  & 0.3s  & 15\%  & 0.5s  & 27\%  \\
    products  & 8.6s    & 4.0s  & 46\%  & 2.2s  & 26\%  & 2.4s  & 28\%  \\
    papers    & 50.4s   & 18.6s & 37\%  & 17.9s & 35\%  & 13.9s & 28\%  \\
    \bottomrule
  \end{tabularx}
  }
  %
  %
  \vskip \floatsep
  %
  %
  %
  \caption{Breakdown of an ogbn-products epoch batch preparation time for \origsystem and \oursystem
    with $P$ threads on 20 cores.
    Note that for \origsystem \emph{Both} column, sampling and slicing occur asynchronously, each using $P$ threads (thus $2P$ in total).
    \oursystem uses only $P$ threads.}
  \tablabel{throughput}
  \vskip 0.05in
  {\small
  \begin{tabularx}{\linewidth}{@{} Xrrrrrr @{}%
    }
    \toprule
    $P$
    & \multicolumn{3}{c}{\emph{PyG}}
    & \multicolumn{3}{c@{}}{\emph{\oursystem}}
    \\
    \cmidrule(lr){2-4}\cmidrule(l){5-7}
    & \multicolumn{1}{c}{\emph{Sampling}}
    & \multicolumn{1}{c}{\emph{Slicing}}
    & \multicolumn{1}{c}{\emph{Both}}
    & \multicolumn{1}{c}{\emph{Sampling}}
    & \multicolumn{1}{c}{\emph{Slicing}}
    & \multicolumn{1}{c@{}}{\emph{Both}}
    \\
    \midrule
     1    & 71.1s   &  7.6s   & 72.7s   & 28.3s   &  7.3s   & 35.6s   \\
    10    & 11.4s   &  1.6s   & 11.5s   &  3.3s   &  0.8s   &  4.1s   \\
    20    &  7.2s   &  1.2s   &  7.3s   &  1.9s   &  0.6s   &  2.5s   \\
    
    \bottomrule
  \end{tabularx}
  }
\end{table}

We benchmarked the per-epoch runtime of the baseline PyG code by 
recording the time required to execute each operation summarized in
\lstref{pytorch-code}.
\punt{Attributing runtime to specific lines of code only tells a
  partial story, of course, but it is still useful for the purpose of
  understanding the apparent performance of each step from a user's
  perspective.}%
Our benchmarks show that batch preparation and CPU-to-GPU data
transfers severely bottleneck training performance.
\tabref{bottleneck} provides a performance breakdown on three publicly
available data sets: ogbn-arxiv, ogbn-products, and ogbn-papers100M
(see \secref{eval} for details).
%
%
%
The reported runtime for each operation is the amount of time
spent on it from the perspective of the main thread executing
the Python code.  In other words, we report the \emph{blocking} time for 
each operation, which is lower than its individual runtime due to 
computation overlap (see \figref{standard-workflow-illustration}).  
Across all three data sets, only about \paperdata{$28$\%} of the time is 
spent on GPU training.  \paperdata{Most of the time} is spent preparing 
batches and transferring data to the GPU.

\punt{Our performance optimizations are principally concerned with
  improving the performance of sampling and data transfers. The
  remainder of this section discusses these two stages of the
  computation in greater depth, and identifies the degree to which
  they bottleneck training time.}

\subsection{Performance analysis of batch preparation}

Batch preparation comprises two steps: (a) neighborhood
\emph{sampling} to obtain the mini-batch induced
subgraph, and (b) \emph{slicing} the feature and label tensors to
extract the parts that correspond to the sampled subgraph. 
Both steps are parallelized: sampling uses 
a PyTorch DataLoader and multiprocessing, and slicing uses multiple 
OpenMP threads in a single process.  The
relative performance of sampling and slicing is not easily obtained
from per-line measurements, as sampling is performed
asynchronously with the main execution thread.  As such, we
investigate the performance of sampling and slicing using separate
targeted benchmarks.

\tabref{throughput} breaks down the performance of sampling and
slicing on ogbn-products for \origsystem.  Batch preparation time is
dominated by the neighborhood sampling time, requiring \paperdata{7.2}
seconds with 20 worker processes.  Slicing, by comparison, takes just
\paperdata{1.2} seconds when parallelized with 20 OpenMP threads using
PyTorch's parallel slicing code.

\punt{These runtimes for sampling and
slicing reflect a best-case scenario where the two parallel operations
do not contend for shared resources.  In practice, the combined
runtime for sampling and slicing in \origsystem is higher and more
variable.}


Even a conservative analysis of the performance breakdown in
\tabreftwo{bottleneck}{throughput} implies that neighborhood sampling is a
substantial bottleneck in GNNs. For \origsystem to
perform sampling at a pace that can keep a single GPU busy and hide sampling 
latency on ogbn-products, sampling throughput must be improved by at least
\paperdata{3$\times$}.  When using multiple GPUs per machine, the required speedup
is higher. \secreftwo{sampler}{shared-memory} discuss how \oursystem
improves the performance of sampling and slicing to alleviate this
bottleneck and achieve substantially higher batch preparation
throughput, as previewed in the \textit{\oursystem} columns of
\tabref{throughput}.

\subsection{Data transfer performance}\seclabel{orig-datatransfer}

Data transfer from CPU to GPU is another bottleneck, accounting for
\paperdata{15--35\%} of the epoch time in the benchmarks of
\tabref{bottleneck}.  Data transfer generally takes longer as
expanded neighborhoods get larger, as seen with ogbn-papers100M, or as
feature dimensionality increases.


There is potential to improve transfer time without also reducing
the amount of transferred data any further.  During a typical epoch with
ogbn-papers100M, a total of \paperdata{164GB} are transferred 
from CPU to GPU\@.  The peak DMA CPU-to-GPU transfer rate on our machine is
\paperdata{12.3GB/s}. 
Per \tabref{bottleneck}, the baseline implementation attains an 
effective data transfer rate of \paperdata{9.2GB/s} or about 
\paperdata{75\%} of peak.
%
%
One can achieve near-optimal data transfer rates with
pipelining and the elimination of redundant round-trip
communications.  These optimizations are discussed in
\secref{datatransfer}.

     \clearpagedraft
\section{\oursystem}
\seclabel{salient}

We propose \oursystem, a system for fast distributed data-parallel GNN
training (and inference; see Section~\ref{sec:inf}) using neighborhood
sampling.  \oursystem combines the following features to achieve high
performance:

\begin{closeenumletters}
\item an optimized implementation of neighborhood sampling and expansion;
\item an efficient parallel batch preparation scheme;
\item CPU-to-GPU data transfer optimizations that hide latency and
  saturate data bus bandwidth; and
\item seamless compatibility with PyTorch's DDP module to scale across
  multiple GPUs and machines.
\end{closeenumletters}

Notably, \oursystem achieves the above without requiring disruptive
changes to user-facing APIs. \oursystem provides a drop-in replacement
for the NeighborSampler and slicing code presently used in \origsystem.

\punt{This section discusses \oursystem's high-performance design and
  analyzes the impact of our key optimizations on alleviating the
  system bottlenecks exposed in \secref{overview}.}

\subsection{Fast neighborhood sampling}
\seclabel{sampler}

\punt{The \texttt{FastMFG} component in \oursystem performs fast
  neighborhood sampling and message-flow graph construction.
  \texttt{FastMFG} achieves a \paperdata{$5\times$} serial performance
  improvement over the \neighborsampler in PyTorch Geometric.  }


The base algorithm for node-wise sampling, implemented in the
NeighborSampler module of \origsystem, is as follows.  We are
given an input graph $G$, a set of nodes $V_{b} = \{v_{1}, \ldots,
v_{k}\}$ which define a mini-batch, and a fanout $d$.  For each node
$v_{i} \in V_{b}$, we sample $d$ of its neighbors without replacement
to get the sampled neighborhood $\mathcal{N}_{d}(v_{i})$.  The sampled
neighborhoods are typically organized into a bipartite graph with
source nodes $\bigcup_{i} \mathcal{N}_{d}(v_{i})$ and destination
nodes $V_{b}$.  For multi-hop neighborhoods, the process is repeated
for each source node, yielding a sequence of bipartite graphs.
Together, these comprise a \textit{message-flow graph}~(MFG) for the
mini-batch of nodes in $V_{b}$.



This simple algorithm for neighborhood sampling admits a variety of
design and implementation choices, which may have a dramatic impact on
performance.  Among the most impactful ones are: a data structure for
global-to-local node ID mapping between the input graph and sampled
MFG; a set data structure to support neighbor sampling without
replacement; and fusing the operations of sampling and MFG
construction.  Overall, the space of possible design choices and
optimizations is too large to explore manually.  We designed a
parameterized implementation of sampled MFG generation to
systematically explore this optimization space and identify the ones
that yield high performance across compute architectures.



\begin{figure}
%
%
\input{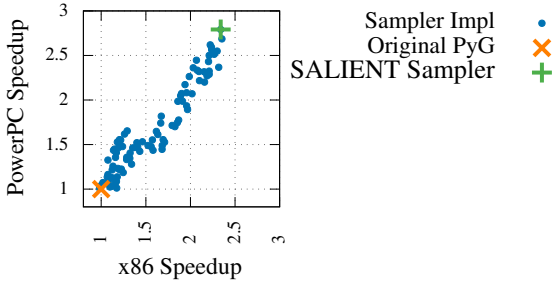}
\vspace{-5ex}
\caption{Exhaustive exploration of optimization parameters.}
\figlabel{data.structure}
\end{figure}

This exploration was done using a microbenchmark which executed the
parameterized code on a reference hop-by-hop trace of the nodes which
made up a sampled MFG for a mini-batch of nodes in ogbn-products.  To
mitigate sampling variability, we benchmark each individual hop of the
reference trace instead of an end-to-end execution.
\figref{data.structure} shows the performance, relative to the
\origsystem NeighborSampler implementation, of $96$
instantiations of the parameterized code on two CPU architectures (x86
and PowerPC).

Analyzing the results shows that the most impactful changes, relative to the
baseline \origsystem code, are related to data structures.
Changing the C++ STL hash map and hash set to a flat swiss-table
implementation \cite{abseil} yields a $2\times$ speedup.
Using an array instead of a hash table for the set provides a further $17\%$
improvement.  Despite its linear search complexity, the array set benefits from
cache locality.  As \tabref{throughput} shows, the \oursystem
implementation of neighborhood sampling is $2.5\times$ faster than 
that of \origsystem.

\subsection{Shared-memory parallel batch preparation}
\seclabel{shared-memory}

\oursystem parallelizes batch preparation through the use of
shared-memory multithreading.  Shared-memory parallelization has several
key advantages over PyTorch's multiprocessing, including lower
synchronization overheads and, critically, the ability to perform
zero-copy communication with the main training process.

%

To parallelize batch preparation across mini-batches, \oursystem uses
C++ threads which prepare batches end-to-end, each
performing sampling and slicing sequentially. 
Since these threads run C++ code, they are not affected by Python's global
interpreter lock.
By using a serial tensor-slicing code, which is otherwise parallelized in
PyTorch by default,
\oursystem improves cache locality and avoids contention between
threads.  Threads balance load dynamically via a lock-free input queue
that contains the destination nodes for each mini-batch.  We find that
dynamic load balancing generally performs better than static
partitioning schemes such as those in the PyTorch DataLoader
due to the variation in final neighborhood size across mini-batches.

A particularly impactful optimization enabled by shared-memory
parallelization is the ability to perform slicing while the main
process is blocked on GPU training.  A batch preparation thread writes
sliced tensors directly into pinned memory accessible by the main process.
By comparison, slicing in PyTorch multiprocessing workers would require
copying the sliced data
from the worker process to the main process via POSIX shared memory,
effectively halving the observed memory bandwidth and inhibiting parallel scaling.

\subsection{Data transfer pipelining}
\seclabel{datatransfer}

Data transfers account for \paperdata{$15$--$35\%$} of per-epoch time as shown
in \tabref{bottleneck}.  To mitigate this bottleneck, \oursystem
employs two optimizations to minimize data transfer latency and
overlap data transfer with GPU computation.


As discussed in \secref{orig-datatransfer}, data transfers for
\origsystem on ogbn-papers100M are only $75\%$ efficient.  Detailed profiling
reveals redundant CPU-GPU round trips which create idle time between
data transfers of the MFG edges.  These round trips are attributed to 
assertions in \origsystem's sparse tensor library
that check the validity of the sparse adjacency matrix after it is transferred.
These blocking assertions are unnecessary for data transfers, since
they have already been performed when the sparse tensor was
constructed on the CPU\@.
Adding an option to skip assertions in such circumstances
allows us to achieve 99\% of peak data transfer throughput.
%

\oursystem further increases GPU utilization by overlapping data
transfers with GPU training computations.  Specifically, \oursystem
uses separate GPU streams for computation and data transfer, 
synchronizing those streams to ensure a training iteration begins
after the necessary data is transferred.
%
%
With \oursystem's optimizations to improve the throughput of batch 
preparation and transfer, these operations generally take less time
than the GPU training computations.  Consequently,
overlapping transfers with GPU computations nearly eliminates latency
outside the GPU computations.

\begin{table}
  \centering
  \caption{Impact of \oursystem optimizations on per-epoch runtime.}
  \small
  \tablabel{summary-of-optimizations}
  \vskip 0.05in
  \begin{tabularx}{\linewidth}{@{} Xrrr @{}}
    \toprule
    \emph{Optimization} & \multicolumn{3}{c@{}}{\emph{Per-Epoch Runtime}} \\
    \cmidrule{2-4}
    & arxiv & products & papers \\
    \midrule
    None (PyG)                  & 1.7s & 8.6s & 50.4s       \\
    + Fast sampling             & 0.7s & 5.3s & 34.6s       \\
    + Shared-memory batch prep. & 0.6s & 4.2s & 27.8s       \\
    + Pipelined data transfers  & 0.5s & 2.8s & 16.5s       \\
    \bottomrule
  \end{tabularx}
\end{table}

\subsection{Summary}
\seclabel{salient-summary}

Our design decisions in \oursystem are informed by a careful analysis
of existing bottlenecks in standard workflows.  We find that it is
possible to get a highly efficient system with
targeted optimizations in neighborhood sampling, shared-memory
parallelization for slicing directly into pinned memory, and pipelining
data transfer and GPU computations.
\figref{salient-workflow-illustration} illustrates the timeline of GNN
training with \oursystem and contrasts it to that of a standard
PyTorch workflow (\figref{standard-workflow-illustration}).
\tabref{summary-of-optimizations} quantifies this comparison,
listing the incremental impact of each optimization category.  These
optimizations do not require fundamental changes to the basic workflow
structure and are orthogonal to other improvements in the training
process itself.
      \clearpagedraft
\section{Inference with Sampling} \label{sec:inf}
While neighborhood sampling is extensively used for training, it is
unclear if this approach compromises prediction accuracy in
inference. Note that these two phases are rather different in
nature. The goal of training is to optimize a loss function and
identify model parameters, whereas the goal of inference is to predict
labels for the test-set nodes. In deep learning, de facto choices of
optimizers are stochastic gradient methods, where the loss function
and the prediction need not be accurately evaluated in every
gradient step to achieve convergence; e.g., the mini-batch gradient is
only an estimator of, but is not exactly, the loss gradient. As long
as sampling is done sufficiently many times, the average will converge
to the probability expectation. Sampling in inference. however, is
one-shot and the sample average may be rather different from the
mean. Will sampling produce predictions as accurate as the case of
non-sampling?

\begin{figure}
\centering
  \input{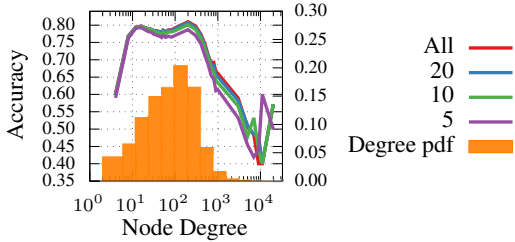}
  \vskip -0.15in
  \caption{Test accuracy and node count versus node degree. Data set:
    ogbn-products; GNN: GraphSAGE. Legend: ``all'' indicates full
    neighborhood (non-sampling); number indicates sampling fanout for
    each layer.  }
  \label{fig:sampling}
\end{figure}

Theoretical analysis is beyond our scope, but we investigate empirical
data. As a typical example, Figure~\ref{fig:sampling} shows the degree
distribution of the test set of ogbn-products overlaid with the
prediction accuracy distribution obtained by using a 3-layer GraphSAGE
architecture. One observes that when the full neighborhood is used,
high-degree nodes tend to be predicted less accurately, but such nodes
are few in the test set. In other words, it suffices to maintain the
prediction quality of the low-degree nodes to achieve a comparable
overall accuracy. Moreover, the figure clearly shows that a small
sampling fanout already approximates well the left half of the
accuracy distribution. As the fanout increases, the right half is
approximated increasingly well, too.

For this reason, we apply neighborhood sampling to inference as
well. It enjoys several advantages. First, it allows reusing the model
architecture code and a majority of the mini-batch training code.
Second, it reduces memory consumption. Because of the explosive size
of multi-hop neighborhoods, a mini-batch is unlikely to fit in GPU memory 
without sampling. Then, inference must be conducted alternatively
by evaluating the network layer by layer and storing layer-wise results
in host memory. For some model architectures (e.g., dense
connections), all layer results must be stored, demanding multiple
times more storage.  Finally, as opposed to the layer-by-layer
approach, mini-batch inference can trivially be run on a select subset of nodes
and can be executed in a distributed data parallel context.


    \clearpagedraft
\section{Evaluation}
\seclabel{eval}

We conduct a comprehensive set of experiments to evaluate the
performance of \oursystem and demonstrate substantial improvement over
a baseline performance-engineered PyG implementation.  All experiments are 
conducted on a cluster of compute nodes in a 10GigE network, each equipped 
with two 20-core Intel Xeon Gold 6248 CPUs, 384GB DRAM, and two NVIDIA 
V100 GPUs (32GB RAM). The benchmarking is based on PyTorch 1.8.1 and PyG
1.7.0. The C++ code for batch preparation is compiled with GCC 7.5.0 and 
optimization flags \texttt{-O3 -march=native}.

\textbf{Data sets.} We evaluate on three standard benchmark data
sets: ogbn-arxiv, ogbn-products, and
ogbn-papers100M~\citep{Hu2020}. The graph and training set in these
data sets vary in size, with ogbn-papers100M being one of the largest
open benchmarks at the time of this work. See Table~\ref{tab:datasets}
for detailed information. All graphs were made undirected (if originally not)
as is common practice. 

\begin{table}
  \caption{Summary of data sets.}
  \label{tab:datasets}
  \vskip 0.05in
  \centering
  \small
  \begin{tabularx}{\linewidth}{@{} Xcccc @{}}
    \toprule
    \emph{Data Set} & \emph{\#Nodes} & \emph{\#Edges} & \emph{\#Feat.} & \emph{Train. / Val. / Test} \\
    \midrule
    arxiv & 169K & 1.2M & 128 & 91K / 30K / 48K \\
    products & 2.4M & 62M & 100 & 197K / 39K / 2.2M \\
    papers & 111M & 1.6B & 128 & 1.2M / 125K / 214K \\
    \bottomrule
  \end{tabularx}
\end{table}

\textbf{GNN architectures.} We experiment with a variety of
architectures to demonstrate the wide applicability of \oursystem:
GraphSAGE~\citep{Hamilton2017}, GAT~\citep{Velickovic2018},
GIN~\citep{Xu2019}, and GraphSAGE-RI. The latter adds residual
connections to GraphSAGE and employs an Inception-like structure for
final prediction.%
\footnote{This architecture is similar to that in the GitHub repo
\href{https://github.com/mengyangniu/ogbn-papers100m-sage}{mengyangniu/ogbn-papers100m-sage}.} 
Details for each GNN atchitecture are given in the appendix. Table~\ref{tab:GNNs}
lists key hyperparameters that impact training time and accuracy. All
experimental results for ogbn-papers100M, except for
Figure~\ref{fig:GNNs}, are obtained with GraphSAGE.

\begin{table}
  \caption{GNN hyperparameters for our experiments. Fanout is for
    training. For inference fanout, see
    Table~\ref{tab:sampling}. Batch size is per GPU.}
  \label{tab:GNNs}
  \vskip 0.05in
  \centering
  \small
  \setlength\tabcolsep{4pt}
  \begin{tabularx}{\linewidth}{@{} XXcccc @{}}
    \toprule
    \emph{Data Set} & \emph{GNN} & \emph{\#Layers} &\emph{Hidden} & \emph{Fanout} & \emph{Batch} \\
    \midrule
    arxiv     & SAGE    & 3 &  256 & (15, 10, 5)  & 1024 \\
    products  & SAGE    & 3 &  256 & (15, 10, 5)  & 1024 \\
    papers    & SAGE    & 3 &  256 & (15, 10, 5)  & 1024 \\
    papers    & GAT     & 3 &  256 & (15, 10, 5)  & 1024 \\
    papers    & GIN     & 3 &  256 & (20, 20, 20) & 1024 \\
    papers    & SAGE-RI & 3 & 1024 & (12, 12, 12) & 1024 \\
    \bottomrule
  \end{tabularx}
\end{table}

\begin{figure}
  \input{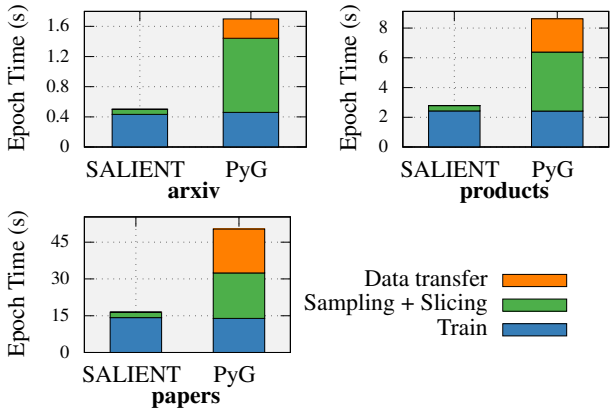}
  \vskip -0.1in
  \caption{Performance improvement of \oursystem over standard PyG
    workflow. Timing measurements on one machine with one GPU.  GNN:
    GraphSAGE with fanout $(15, 10, 5)$.}
  \label{fig:perf.engineer}
\end{figure}

\textbf{Single-GPU improvement over PyG.} We first compare the
performance of \oursystem and PyG on a single
GPU\@. Figure~\ref{fig:perf.engineer} suggests a
\paperdata{3$\times$} to \paperdata{3.4$\times$} speedup across data
sets, owing to the diminishing percentage of time blocked on sampling
and data transfer. \oursystem's optimizations improve the overall
efficiency of these stages, and its pipelined design results in the overall per-epoch
runtime being nearly equal to the GPU compute time for training.

\punt{by removing unnecessary round-trip
CPU-to-GPU communication during transfers, avoiding unnecessary data copying by using shared-memory parallelism, and reducing sampling time with carefully tuned codes. These improvements, combined with \oursystem's pipelined design, result in
the overall per-epoch runtime being nearly equal to the GPU compute
time for training.}

\begin{figure*}
  \input{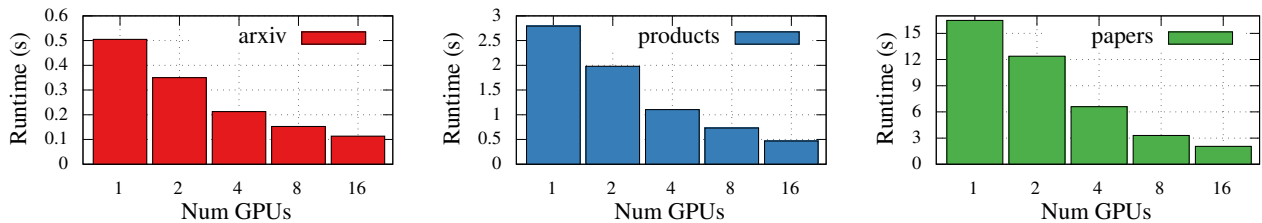}
  \vskip -0.25in
	\caption{Epoch time when scaling to multiple GPUs with
      proportionately scaled batch size using the SAGE architecture
      from \tabref{GNNs}.  }
  \label{fig:scaling}
\end{figure*}

\begin{figure}
  \centering
  \vskip -0.1in
	\input{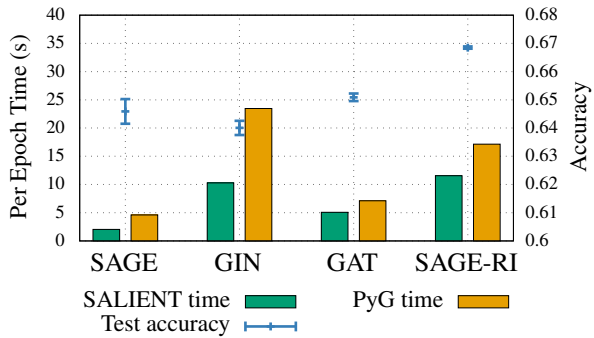}
  \vskip -0.25in
	\caption{Per epoch training time and test accuracy after 25 epochs
      for ogbn-papers100M on several GNN models, trained by using 16
      GPUs. Test inference fanouts were $(20, 20, 20)$ for SAGE, GIN,
      and GAT, and $(100, 100, 100)$ for SAGE-RI.}
  \label{fig:GNNs}
\end{figure}

\textbf{Multi-GPU scaling.} We now scale the training to multiple
GPUs.  A maximum of 16 GPUs are used, spanning eight machines. The
effective batch size is proportional to the number of GPUs.
\oursystem straightforwardly applies the PyTorch DDP module and
performs distributed communications with the NCCL backend.
Figure~\ref{fig:scaling} shows generally good scaling in the
distributed setting. Larger data sets, such as ogbn-papers100M, tend
to see greater parallel speedup due to having higher computational
density and larger training sets. As such, bigger graphs amortize the
latency of starting an epoch (e.g., the time to prepare the first sets
of mini-batches) over a greater amount of work per GPU. The sampled
neighborhoods of batches also tend to be larger for bigger and
well-connected graphs, which increases the amount of GPU computations
per mini-batch and better shadows communication and synchronization
overheads. With 16 GPUs, the speedup ranges from
\paperdata{4.45$\times$} to \paperdata{8.05$\times$}.

\textbf{Neighborhood sampling for inference.} We investigate the
effectiveness of applying neighborhood sampling for
inference. Table~\ref{tab:sampling} lists the test accuracies for all
data sets either using either full or sampled neighborhoods. Each
accuracy result is obtained through five repetitions of training and
inference.  Full-neighborhood inference uses layer-wise computation
and stores intermediate layer results in host memory. One observes
that for the ogbn-arxiv and ogbn-products data sets, a fanout of 20
for each layer is sufficient to match full-neighborhood accuracy. For ogbn-papers100M, layer-wise inference with full neighborhood
runs out of memory.  Hence, we report the accuracy with fanout
\paperdata{100} instead. We see that the accuracy has been
saturated and conclude that fanout 20 is sufficient for this data
set as well.

\begin{table}
  \caption{Test accuracy under various neighborhood fanouts for
    inference. GNN: GraphSAGE with training fanout (15, 10, 5). For
    ogbn-papers100M, the ``fanout: all'' case runs out of memory and
    we report the result with fanout (100, 100, 100) instead.}
  \label{tab:sampling}
  \vskip 0.05in
  \centering
  \footnotesize
  \tabcolsep 3.9pt
  \begin{tabularx}{\linewidth}{@{} Xcccc @{}}
    \toprule
    \emph{Data Set} & \multicolumn{4}{c@{}}{\emph{Accuracy}} \\
    \cmidrule(l){2-5}
    & \emph{fanout: all} & \emph{(20, 20, 20)} & \emph{(10, 10, 10)} & \emph{(5, 5, 5)} \\
    \midrule
    arxiv & .7074$\pm$.005 & .7054$\pm$.005 & .6980$\pm$.005 & .6849$\pm$.004 \\
    products & .7749$\pm$.004 & .7755$\pm$.003 & .7708$\pm$.003 & .7558$\pm$.003\\
	papers & \paperdata{.6491$\pm$.005$^*\!\!\!$} & .6458$\pm$.004 & .6379$\pm$.004 & .6163$\pm$.005\\
    \bottomrule
  \end{tabularx}
\end{table}

\textbf{Performance of varying GNNs.} A feature of \oursystem is that
the GNN architecture implementation is independent of performance
engineering in batch preparation and transfer.  Hence, a PyG developer 
can keep using exactly the same API to design and tune GNNs.  This
feature brings in the benefit of fast prototyping for an
application.  We experiment with a number of architectures and report
the training time (with 16 GPUs) and test accuracy for the largest
data set ogbn-papers100M in Figure~\ref{fig:GNNs}. 

Several observations follow. First, the training time for different
architectures varies significantly, affected by multiple factors
such as the complexity of the architecture and the choice of
hyperparameters.  Second, speedup over PyG also varies
significantly. Computation density (relative to data transfer
and communication) is highest for GraphSAGE-RI, medium for GAT and
GIN, and lowest for GraphSAGE. GraphSAGE enjoys the greatest
improvement (approximately \paperdata{2.3$\times$}) due to our
performance engineering on sampling and transfer, while GraphSAGE-RI and
GAT have the least improvement, which however is still over
\paperdata{1.4$\times$}.  Third, architectures achieve different
accuracies. With only moderate tuning, GraphSAGE-RI performs
noticeably better than the other three. These accuracy numbers are on
par with those appearing in the literature or public GitHub repos.
         \clearpagedraft
\section{Comparison with Existing Systems}
\label{sec:comparison}

It is important to put the results in \secref{eval} in
perspective. Table~\ref{tab:comparison} summarizes the reported
performance of several representative systems. On the largest data
set, ogbn-papers100M, our \paperdata{2.0s} per-epoch training time is
orders of magnitude faster than that of the listed systems. This
record, however, is achieved in an incomparable environment (differing
in hardware, software framework, model architecture, or batching
scheme) from those of existing systems. Since most referenced systems
are not publicly available or readily usable, we note a few
differentiating points.

{\renewcommand{\arraystretch}{1.5}
  \begin{table*}[!t] 
    \centering
    \caption{Representative GNN training systems and their performance on either ogbn-papers100M or the largest graph reported, whichever is larger, for each system.}
    \label{tab:comparison}
    \vskip 0.05in
    \begin{threeparttable}
      \centering
      \small
      \tabcolsep 4.5pt
      \begin{tabularx}{\linewidth}{%
        @{} >{\columncolor{white}[0pt][\tabcolsep]}
        l                                                  
        p{\widthof{\textbf{Framework}}}                    
        p{\widthof{$d^{\ell} \!=\! (\text{cache},15,10)$}} 
        p{\widthof{GraphSAGE,}}                            
        L                                                  
        p{\widthof{ogbn-papers100M:}}                      
        p{\widthof{Infer: 1.74}}                           
        >{\columncolor{white}[\tabcolsep][0pt]}
        p{\widthof{63.31 \textsuperscript{f}}}             
        @{}
        }
        \toprule
        \textbf{System} &
        \textbf{Framework} &
        \textbf{Batching} &
        \textbf{GNN} &
        \textbf{Machines} &
        \textbf{Data Set} &
        \textbf{Speed} (s/epoch) &
        \textbf{Acc.}\newline(\%) \\
        \midrule
        \arrayrulecolor{gray}
        NeuGraph &
        TensorFlow &
        full-batch & 
        GCN, \newline $L = 2$ &
        1 machine with \newline 28 Intel cores, \newline 512GB DRAM, \newline 8 P100 GPUs &
        amazon: \newline $|V| = 8.6$M, \newline $|E| = 231.6$M, \newline $f = 96$ \newline {\scriptsize \citep{McAuley2015}} &
        0.655 \tnote{a} &
        N/A \\
        \hline
        Roc &
        FlexFlow, Lux &
        full-batch & 
        GCN &
        4 machines, each has \newline 20 x86 cores, \newline 256GB DRAM, \newline 4 P100 GPUs; \newline 100Gbps InfiniBand &
        amazon: \newline $|V| = 9.4$M, \newline $|E| = 231.6$M, \newline $f = 300$ \newline {\scriptsize \citep{He2016}} &
        0.526 \tnote{b} &
        N/A \\
        \hline
        DistDGL &
        PyTorch, DGL, METIS &
        mini-batch, \newline size 2000, \newline $d^{\ell} \!=\! (15, 10, 5)$ & 
        GraphSAGE, \newline $L = 3$, \newline $f_{\text{hidden}} = 256$  &
        16 EC2 instances, each has 96 vCPUs, \newline 384GB DRAM; \newline 100Gbps network & 
        ogbn-papers100M: \newline $|V| = 111$M, \newline $|E| = 1.6$B, \newline $f = 128$ \newline {\scriptsize \citep{Hu2020}} &
        13 \tnote{c} &
        N/A \\
        \hline
        DeepGalois &
        Galois, GuSP, Gluon &
        full-batch & 
        GraphSAGE, \newline $L = 2$, \newline $f_{\text{hidden}} = 16$ &
        32 machines, each has 48 x86 cores, \newline 192GB DRAM; \newline 100Gbps Omni-Path & 
        same as above &
        70 \tnote{d} &
        N/A \\
        \hline
        Zero-Copy &
        PyTorch, DGL &
        mini-batch & 
        GraphSAGE &
        1 machine with \newline 24 AMD cores, \newline 256GB DRAM, \newline 2 RTX3090 GPUs &
        same as above &
        648 \tnote{e} &
        N/A \\
        \hline
        GNS &
        PyTorch, DGL &
        mini-batch, \newline size 1000, \newline $d^{\ell} \!=\! (\text{cache},15,10)$ & 
        GraphSAGE, \newline $L = 3$, \newline $f_{\text{hidden}} = 256$ &
        1 EC2 instance with \newline 32 CPU cores, \newline 256GB DRAM, \newline 1 T4 GPU  &
        same as above &
        98.5 \tnote{f} &
        63.31 \tnote{f} \\
        \hline
        $P^3$ &
        PyTorch, DGL &
        mini-batch, \newline size 1000, \newline $d^{\ell} \!=\! (25, 10)$ &
        GraphSAGE, \newline $L=2$, \newline $f_{\text{hidden}} = 32$ &
        4 machines, each has \newline 1$\times$12-core Intel CPUs, \newline 441GB DRAM, \newline 4 P100 GPUs; \newline 10Gbps Ethernet &
        same as above &
        3.107 \tnote{g} &
        N/A \\
        \hline
        \rowcolor{Gray}
        \oursystem &
        PyTorch, PyG, \newline DDP &
        mini-batch, \newline size \paperdata{1024}, \newline
        $d^{\ell}_{\text{train}} \!=\! (15, 10, 5)$, \newline
        $d^{\ell}_{\text{infer}} \!=\! (20, 20, 20)$ & 
        GraphSAGE, \newline $L = 3$, \newline $f_{\text{hidden}} = 256$ &
        \paperdata{%
        8 machines, each has \newline 2$\times$20-core Intel CPUs, \newline 384GB DRAM, \newline 2 V100 GPUs; \newline 10GigE network} &
        same as above &
        Train: \paperdata{2.0} \newline\newline Infer:~\paperdata{2.4}s \newline 
        on test set &
        64.58 $\pm$0.40 \\
        \arrayrulecolor{black}
        \bottomrule
      \end{tabularx}
      \begin{tablenotes}
      \item [a] Estimated as 6.55/10, where 6.55 comes from the TF-SAGA section of Table 2 and 10 is estimated from Figure 17 of \citet{Ma2019}.
      \item [b] Estimated as 1/1.9, where 1.9 is estimated from Figure 5 of \citet{Jia2020}.
      \item [c] Reported in Figure 8 of \citet{Zheng2020}.
      \item [d] Estimated from Figure 4 of \citet{Hoang2021}. Note that the referenced article demonstrates that under the same full-batch setting, DeepGalois may be several times faster than DistDGL.
      \item [e] Estimated from Figure 11 of \citet{Min2021}.
      \item [f] Reported in Table 3 of \citet{Dong2021}.
      \item [g] Reported in Table 4 of \citet{Gandhi2021}.
      \end{tablenotes}
    \end{threeparttable}
\end{table*}}

We adopt mini-batch training, as opposed to full-batch training
appearing in several prior systems. One reason is that the former
converges faster and generalizes better~\citep{Bottou2018}. On the
system level, these two batching schemes have drastically different
computation patterns and may suffer different bottlenecks.
%
%

\oursystem is built on PyTorch and PyG, a framework less used in system-oriented publications. We consider that PyG enjoys a large user base%
\footnote{At the time of this writing,
\href{https://github.com/pyg-team/pytorch_geometric}{pyg-team/pytorch\_geometric}
has 12.8K stars and 2.2K forks on GitHub, while
\href{https://github.com/dmlc/dgl}{dmlc/dgl} has 8.2K stars and 1.8K
forks and \href{https://github.com/alibaba/euler}{alibaba/euler} has
2.7K stars and 534 forks.}  and it benefits from a demonstration of
improvement that encourages widespread attraction. Meanwhile, it
should be noted that \oursystem's optimizations are general and can be
applied to other frameworks.
%

We demonstrate experiments in a multi-machine, multi-GPU environment,
with attractive speedup, using PyTorch's DDP module for distributed
training. Most of the systems summarized in
Table~\ref{tab:comparison} demonstrate no such experiments and/or are
not readily extensible to such an environment. The
  only exception, adopting mini-batch training, is
  $P^3$~\citep{Gandhi2021}. This system addresses a different
  bottleneck than we do---the communication cost and partitioning
  overhead. The techniques proposed therein are independent of
  \oursystem and can be incorporated into our sampling pipeline for a
  further efficient system.



   \clearpagedraft
\section{Conclusions and Future Work}

In this work, we identify major bottlenecks in GNN training and
inference---batch preparation and transfer---and propose three
complementary improvements, namely optimized neighborhood sampling,
shared-memory parallel sampling and slicing, and pipelined data transfers. 
We also find that neighborhood sampling impacts 
inference accuracy only minimally.  We build our system \oursystem 
based on PyTorch and PyG and showcase that changing the GNN 
architecture can be easily done as usual, without interfering with the
training/inference code.


We demonstrate that \oursystem achieves near-perfect overlap of batch
preparation, transfer, and training computations.  That is, the 
end-to-end training time per epoch is nearly equal to the time for the
slowest of these components in isolation.  The limiting factor for batch
preparation is the number of CPU cores or the DRAM bandwidth; for data 
transfer it is the peak CPU-to-GPU memory bandwidth.  As feature vector 
size increases, or with higher fanout, memory bandwidth may become 
insufficient.  Then, one must avail of additional techniques such 
as GPU-based slicing~\citep{Min2021} or caching data on the 
GPU~\citep{Dong2021} to reduce the slicing or data transfer volume.  

An additional avenue of future work is distributing the graph and node
data in a distributed computing environment to accommodate processing
even larger graphs.  Graph partitioning~\citep{Karypis1999} will
likely be invoked, but the objective may consider not only edge cut
and load balance but also the cost of multi-hop neighborhood sampling.
Sampling approaches will need to be re-investigated in a distributed
environment, to minimize communication. Partitioning
  along the feature dimension is another promising technique for long
  feature vectors~\citep{Gandhi2021}.

\section*{Acknowledgements}
This work was conducted on the SuperCloud computing cluster
\url{https://supercloud.mit.edu} and the Satori computing cluster
\url{https://mit-satori.github.io}. This research was sponsored by
MIT-IBM Watson AI Lab and in part by the United States Air Force
Research Laboratory and the United States Air Force Artificial
Intelligence Accelerator and was accomplished under Cooperative
Agreement Number FA8750-19-2-1000.  The views and conclusions
contained in this document are those of the authors and should not be
interpreted as representing the official policies, either expressed or
implied, of the United States Air Force or the U.S. Government.  The
U.S. Government is authorized to reproduce and distribute reprints for
Government purposes notwithstanding any copyright notation herein.

\bibliography{reference}

\begin{thebibliography}{43}
\providecommand{\natexlab}[1]{#1}
\providecommand{\url}[1]{\texttt{#1}}
\expandafter\ifx\csname urlstyle\endcsname\relax
  \providecommand{\doi}[1]{doi: #1}\else
  \providecommand{\doi}{doi: \begingroup \urlstyle{rm}\Url}\fi

\bibitem[Abadi et~al.(2015)Abadi, Agarwal, Barham, Brevdo, Chen, Citro,
  Corrado, Davis, Dean, Devin, Ghemawat, Goodfellow, Harp, Irving, Isard, Jia,
  Jozefowicz, Kaiser, Kudlur, Levenberg, Man\'{e}, Monga, Moore, Murray, Olah,
  Schuster, Shlens, Steiner, Sutskever, Talwar, Tucker, Vanhoucke, Vasudevan,
  Vi\'{e}gas, Vinyals, Warden, Wattenberg, Wicke, Yu, and Zheng]{Abadi2015}
Abadi, M., Agarwal, A., Barham, P., Brevdo, E., Chen, Z., Citro, C., Corrado,
  G.~S., Davis, A., Dean, J., Devin, M., Ghemawat, S., Goodfellow, I., Harp,
  A., Irving, G., Isard, M., Jia, Y., Jozefowicz, R., Kaiser, L., Kudlur, M.,
  Levenberg, J., Man\'{e}, D., Monga, R., Moore, S., Murray, D., Olah, C.,
  Schuster, M., Shlens, J., Steiner, B., Sutskever, I., Talwar, K., Tucker, P.,
  Vanhoucke, V., Vasudevan, V., Vi\'{e}gas, F., Vinyals, O., Warden, P.,
  Wattenberg, M., Wicke, M., Yu, Y., and Zheng, X.
\newblock {TensorFlow}: Large-scale machine learning on heterogeneous systems,
  2015.
\newblock https://www.tensorflow.org/.

\bibitem[Benzaquen et~al.(2018)Benzaquen, Evlogimenos, Kulkunidis, and
  Pereplitsa]{abseil}
Benzaquen, S., Evlogimenos, A., Kulkunidis, M., and Pereplitsa, R.
\newblock Swiss tables and absl::hash, Sep 2018.
\newblock URL \url{https://abseil.io/blog/20180927-swisstables}.

\bibitem[Bottou et~al.(2018)Bottou, Curtis, and Nocedal]{Bottou2018}
Bottou, L., Curtis, F.~E., and Nocedal, J.
\newblock Optimization methods for large-scale machine learning.
\newblock \emph{SIAM Rev.}, 60\penalty0 (2):\penalty0 223--311, 2018.

\bibitem[Chen \& Luss(2018)Chen and Luss]{Chen2018a}
Chen, J. and Luss, R.
\newblock Stochastic gradient descent with biased but consistent gradient
  estimators.
\newblock Preprint arXiv:1807.11880, 2018.

\bibitem[Chen et~al.(2018)Chen, Ma, and Xiao]{Chen2018}
Chen, J., Ma, T., and Xiao, C.
\newblock {FastGCN}: Fast learning with graph convolutional networks via
  importance sampling.
\newblock In \emph{ICLR}, 2018.

\bibitem[Chiang et~al.(2019)Chiang, Liu, Si, Li, Bengio, and Hsieh]{Chiang2019}
Chiang, W.-L., Liu, X., Si, S., Li, Y., Bengio, S., and Hsieh, C.-J.
\newblock Cluster-{GCN}: An efficient algorithm for training deep and large
  graph convolutional networks.
\newblock In \emph{KDD}, 2019.

\bibitem[Dong et~al.(2021)Dong, Zheng, Yang, and Karypis]{Dong2021}
Dong, J., Zheng, D., Yang, L.~F., and Karypis, G.
\newblock Global neighbor sampling for mixed {CPU-GPU} training on giant
  graphs.
\newblock In \emph{KDD}, 2021.

\bibitem[Fey \& Lenssen(2019)Fey and Lenssen]{Fey2019}
Fey, M. and Lenssen, J.~E.
\newblock Fast graph representation learning with {PyTorch Geometric}.
\newblock In \emph{ICLR Workshop on Representation Learning on Graphs and
  Manifolds}, 2019.

\bibitem[Gandhi \& Iyer(2021)Gandhi and Iyer]{Gandhi2021}
Gandhi, S. and Iyer, A.~P.
\newblock {P3}: Distributed deep graph learning at scale.
\newblock In \emph{OSDI}, 2021.

\bibitem[Gilmer et~al.(2017)Gilmer, Schoenholz, Riley, Vinyals, and
  Dahl]{Gilmer2017}
Gilmer, J., Schoenholz, S.~S., Riley, P.~F., Vinyals, O., and Dahl, G.~E.
\newblock Neural message passing for quantum chemistry.
\newblock In \emph{ICML}, 2017.

\bibitem[Hamilton et~al.(2017)Hamilton, Ying, and Leskovec]{Hamilton2017}
Hamilton, W.~L., Ying, R., and Leskovec, J.
\newblock Inductive representation learning on large graphs.
\newblock In \emph{NIPS}, 2017.

\bibitem[He \& McAuley(2016)He and McAuley]{He2016}
He, R. and McAuley, J.
\newblock Ups and downs: Modeling the visual evolution of fashion trends with
  one-class collaborative filtering.
\newblock In \emph{WWW}, 2016.

\bibitem[Hoang et~al.(2021)Hoang, Chen, Lee, Dathathri, Gill, and
  Pingali]{Hoang2021}
Hoang, L., Chen, X., Lee, H., Dathathri, R., Gill, G., and Pingali, K.
\newblock Efficient distribution for deep learning on large graphs,.
\newblock In \emph{GNNSys}, 2021.

\bibitem[Hu et~al.(2020{\natexlab{a}})Hu, Fey, Zitnik, Dong, Ren, Liu, Catasta,
  and Leskovec]{Hu2020}
Hu, W., Fey, M., Zitnik, M., Dong, Y., Ren, H., Liu, B., Catasta, M., and
  Leskovec, J.
\newblock Open graph benchmark: Datasets for machine learning on graphs.
\newblock Preprint arXiv:2005.00687, 2020{\natexlab{a}}.

\bibitem[Hu et~al.(2020{\natexlab{b}})Hu, Liu, Gomes, Zitnik, Liang, Pande, and
  Leskovec]{Hu2020a}
Hu, W., Liu, B., Gomes, J., Zitnik, M., Liang, P., Pande, V., and Leskovec, J.
\newblock Strategies for pre-training graph neural networks.
\newblock In \emph{ICLR}, 2020{\natexlab{b}}.

\bibitem[Jia et~al.(2020)Jia, Lin, Gao, Zaharia, and Aiken]{Jia2020}
Jia, Z., Lin, S., Gao, M., Zaharia, M., and Aiken, A.
\newblock Improving the accuracy, scalability, and performance of graph neural
  networks with {Roc}.
\newblock In \emph{MLSys}, 2020.

\bibitem[Karypis \& Kumar(1999)Karypis and Kumar]{Karypis1999}
Karypis, G. and Kumar, V.
\newblock A fast and high quality multilevel scheme for partitioning irregular
  graphs.
\newblock \emph{SIAM Journal on Scientific Computing}, 20\penalty0
  (1):\penalty0 359--392, 1999.

\bibitem[Kingma \& Ba(2015)Kingma and Ba]{Kingma2015}
Kingma, D.~P. and Ba, J.
\newblock Adam: A method for stochastic optimization.
\newblock In \emph{ICLR}, 2015.

\bibitem[Kipf \& Welling(2017)Kipf and Welling]{Kipf2017}
Kipf, T.~N. and Welling, M.
\newblock Semi-supervised classification with graph convolutional networks.
\newblock In \emph{ICLR}, 2017.

\bibitem[Lawson et~al.(1979)Lawson, Hanson, Kincaid, , and Krogh]{Lawson1979}
Lawson, C.~L., Hanson, R.~J., Kincaid, D., , and Krogh, F.~T.
\newblock Basic linear algebra subprograms for {FORTRAN} usage.
\newblock \emph{ACM Trans. Math. Soft.}, 5:\penalty0 308--323, 1979.

\bibitem[Li et~al.(2016)Li, Tarlow, Brockschmidt, and Zemel]{Li2016}
Li, Y., Tarlow, D., Brockschmidt, M., and Zemel, R.
\newblock Gated graph sequence neural networks.
\newblock In \emph{ICLR}, 2016.

\bibitem[Li et~al.(2018)Li, Yu, Shahabi, and Liu]{Li2018}
Li, Y., Yu, R., Shahabi, C., and Liu, Y.
\newblock Diffusion convolutional recurrent neural network: Data-driven traffic
  forecasting.
\newblock In \emph{ICLR}, 2018.

\bibitem[Ma et~al.(2019)Ma, Yang, Miao, Xue, Wu, Zhou, and Dai]{Ma2019}
Ma, L., Yang, Z., Miao, Y., Xue, J., Wu, M., Zhou, L., and Dai, Y.
\newblock {NeuGraph}: Parallel deep neural network computation on large graphs.
\newblock In \emph{USENIX ATC}, 2019.

\bibitem[Ma \& Chen(2021)Ma and Chen]{Ma2021}
Ma, T. and Chen, J.
\newblock Unsupervised learning of graph hierarchical abstractions with
  differentiable coarsening and optimal transport.
\newblock In \emph{AAAI}, 2021.

\bibitem[McAuley et~al.(2015)McAuley, Targett, Shi, and van~den
  Hengel]{McAuley2015}
McAuley, J., Targett, C., Shi, Q., and van~den Hengel, A.
\newblock Image-based recommendations on styles and substitutes.
\newblock In \emph{SIGIR}, 2015.

\bibitem[Min et~al.(2021)Min, Wu, Huang, Hidayetoğlu, Xiong, Ebrahimi, Chen,
  and mei Hwu]{Min2021}
Min, S.~W., Wu, K., Huang, S., Hidayetoğlu, M., Xiong, J., Ebrahimi, E., Chen,
  D., and mei Hwu, W.
\newblock Large graph convolutional network training with {GPU}-oriented data
  communication architecture.
\newblock Preprint arXiv:2103.03330, 2021.

\bibitem[Mirhoseini et~al.(2021)Mirhoseini, Goldie, Yazgan, Jiang, Songhori,
  Wang, Lee, Johnson, Pathak, Nazi, Pak, Tong, Srinivasa, Hang, Tuncer, Le,
  Laudon, Ho, Carpenter, and Dean]{Mirhoseini2021}
Mirhoseini, A., Goldie, A., Yazgan, M., Jiang, J.~W., Songhori, E., Wang, S.,
  Lee, Y.-J., Johnson, E., Pathak, O., Nazi, A., Pak, J., Tong, A., Srinivasa,
  K., Hang, W., Tuncer, E., Le, Q.~V., Laudon, J., Ho, R., Carpenter, R., and
  Dean, J.
\newblock A graph placement methodology for fast chip design.
\newblock \emph{Nature}, 594:\penalty0 207--212, 2021.

\bibitem[Morris et~al.(2019)Morris, Ritzert, Fey, Hamilton, Lenssen, Rattan,
  and Grohe]{Morris2019}
Morris, C., Ritzert, M., Fey, M., Hamilton, W.~L., Lenssen, J.~E., Rattan, G.,
  and Grohe, M.
\newblock {W}eisfeiler and {L}eman go neural: Higher-order graph neural
  networks.
\newblock In \emph{AAAI}, 2019.

\bibitem[Paszke et~al.(2017)Paszke, Gross, Chintala, Chanan, Yang, DeVito, Lin,
  Desmaison, Antiga, and Lerer]{Paszke2017}
Paszke, A., Gross, S., Chintala, S., Chanan, G., Yang, E., DeVito, Z., Lin, Z.,
  Desmaison, A., Antiga, L., and Lerer, A.
\newblock Automatic differentiation in {PyTorch}.
\newblock In \emph{NIPS 2017 Autodiff Workshop}, 2017.

\bibitem[Paszke et~al.(2019)Paszke, Gross, Massa, Lerer, Bradbury, Chanan,
  Killeen, Lin, Gimelshein, Antiga, Desmaison, Kopf, Yang, DeVito, Raison,
  Tejani, Chilamkurthy, Steiner, Fang, Bai, and Chintala]{Paszke2019}
Paszke, A., Gross, S., Massa, F., Lerer, A., Bradbury, J., Chanan, G., Killeen,
  T., Lin, Z., Gimelshein, N., Antiga, L., Desmaison, A., Kopf, A., Yang, E.,
  DeVito, Z., Raison, M., Tejani, A., Chilamkurthy, S., Steiner, B., Fang, L.,
  Bai, J., and Chintala, S.
\newblock {PyTorch}: An imperative style, high-performance deep learning
  library.
\newblock In \emph{NIPS}, 2019.

\bibitem[Ramezani et~al.(2020)Ramezani, Cong, Mahdavi, Sivasubramaniam, and
  Kandemir]{Ramezani2020}
Ramezani, M., Cong, W., Mahdavi, M., Sivasubramaniam, A., and Kandemir, M.
\newblock {GCN} meets {GPU}: Decoupling ``when to sample'' from ``how to
  sample''.
\newblock In \emph{NeurIPS}, 2020.

\bibitem[Shang et~al.(2021)Shang, Chen, and Bi]{Shang2021}
Shang, C., Chen, J., and Bi, J.
\newblock Discrete graph structure learning for forecasting multiple time
  series.
\newblock In \emph{ICLR}, 2021.

\bibitem[Veli\v{c}kovi\'{c} et~al.(2018)Veli\v{c}kovi\'{c}, Cucurull, Casanova,
  Romero, Li\`{o}, and Bengio]{Velickovic2018}
Veli\v{c}kovi\'{c}, P., Cucurull, G., Casanova, A., Romero, A., Li\`{o}, P.,
  and Bengio, Y.
\newblock Graph attention networks.
\newblock In \emph{ICLR}, 2018.

\bibitem[Wang et~al.(2021{\natexlab{a}})Wang, Yin, Tian, Yang, Chen, Yu, Yao,
  and Zhou]{Wang2021}
Wang, L., Yin, Q., Tian, C., Yang, J., Chen, R., Yu, W., Yao, Z., and Zhou, J.
\newblock {FlexGraph}: a flexible and efficient distributed framework for {GNN}
  training.
\newblock In \emph{EuroSys}, 2021{\natexlab{a}}.

\bibitem[Wang et~al.(2019)Wang, Zheng, Ye, Gan, Li, Song, Zhou, Ma, Yu, Gai,
  Xiao, He, Karypis, Li, and Zhang]{Wang2019}
Wang, M., Zheng, D., Ye, Z., Gan, Q., Li, M., Song, X., Zhou, J., Ma, C., Yu,
  L., Gai, Y., Xiao, T., He, T., Karypis, G., Li, J., and Zhang, Z.
\newblock Deep graph library: A graph-centric, highly-performant package for
  graph neural networks.
\newblock Preprint arXiv:1909.01315, 2019.

\bibitem[Wang et~al.(2021{\natexlab{b}})Wang, Feng, Li, Li, Deng, Xie, and
  Ding]{Wang2021a}
Wang, Y., Feng, B., Li, G., Li, S., Deng, L., Xie, Y., and Ding, Y.
\newblock {GNNAdvisor}: An adaptive and efficient runtime system for {GNN}
  acceleration on {GPUs}.
\newblock In \emph{OSDI}, 2021{\natexlab{b}}.

\bibitem[Weber et~al.(2019)Weber, Domeniconi, Chen, Weidele, Bellei, Robinson,
  and Leiserson]{Weber2019}
Weber, M., Domeniconi, G., Chen, J., Weidele, D. K.~I., Bellei, C., Robinson,
  T., and Leiserson, C.~E.
\newblock Anti-money laundering in {B}itcoin: Experimenting with graph
  convolutional networks for financial forensics.
\newblock In \emph{KDD Workshop on Anomaly Detection in Finance}, 2019.

\bibitem[Wu et~al.(2021)Wu, Ma, Cai, Jin, Li, Zheng, Cheng, and Yu]{Wu2021}
Wu, Y., Ma, K., Cai, Z., Jin, T., Li, B., Zheng, C., Cheng, J., and Yu, F.
\newblock {Seastar}: vertex-centric programming for graph neural networks.
\newblock In \emph{EuroSys}, 2021.

\bibitem[Xu et~al.(2019)Xu, Hu, Leskovec, and Jegelka]{Xu2019}
Xu, K., Hu, W., Leskovec, J., and Jegelka, S.
\newblock How powerful are graph neural networks?
\newblock In \emph{ICLR}, 2019.

\bibitem[Ying et~al.(2018)Ying, He, Chen, Eksombatchai, Hamilton, and
  Leskovec]{Ying2018}
Ying, R., He, R., Chen, K., Eksombatchai, P., Hamilton, W.~L., and Leskovec, J.
\newblock Graph convolutional neural networks for web-scale recommender
  systems.
\newblock In \emph{KDD}, 2018.

\bibitem[Zeng et~al.(2020)Zeng, Zhou, Srivastava, Kannan, and
  Prasanna]{Zeng2020}
Zeng, H., Zhou, H., Srivastava, A., Kannan, R., and Prasanna, V.
\newblock {GraphSAINT}: Graph sampling based inductive learning method.
\newblock In \emph{ICLR}, 2020.

\bibitem[Zheng et~al.(2020)Zheng, Ma, Wang, Zhou, Su, Song, Gan, Zhang, and
  Karypis]{Zheng2020}
Zheng, D., Ma, C., Wang, M., Zhou, J., Su, Q., Song, X., Gan, Q., Zhang, Z.,
  and Karypis, G.
\newblock {{DistDGL}}: Distributed graph neural network training for
  billion-scale graphs.
\newblock In \emph{IA3}, 2020.

\bibitem[Zou et~al.(2019)Zou, Hu, Wang, Jiang, Sun, and Gu]{Zou2019}
Zou, D., Hu, Z., Wang, Y., Jiang, S., Sun, Y., and Gu, Q.
\newblock Layer-dependent importance sampling for training deep and large graph
  convolutional networks.
\newblock In \emph{NeurIPS}, 2019.

\end{thebibliography}
\bibliographystyle{mlsys2022}

\newpage
\appendix


\section{Artifact Appendix}

\subsection{Abstract}
 
This section describes the software artifacts associated with this paper for the purpose of replicating the presented experimental results. The code is distributed via GitHub at \url{https://github.com/MITIBMxGraph/SALIENT_artifact} and can be
used to perform the experiments presented in the paper. To streamline the process of exercising
the software to reproduce key experimental results, we have provided scripts in the \texttt{experiments}
directory of the repository to run: (a) single GPU experiments that produce data for \tabref{bottleneck} and Figure~\ref{fig:perf.engineer};
and, (b) distributed multi-GPU experiments that produce data for Figure~\ref{fig:scaling} and Figure~\ref{fig:GNNs}. Detailed instructions for running these scripts are provided in a dedicated readme file for artifact evaluation located at \url{https://github.com/MITIBMxGraph/SALIENT_artifact/blob/main/README.md}. 



\subsection{Artifact check-list (meta-information)}


\begin{itemize}[leftmargin=*]
  \item \textbf{Algorithm:} PyG and \oursystem training algorithms for GNNs with node-wise sampling.
  \item \textbf{Program:} PyTorch, CUDA
  \item \textbf{Compilation:} \texttt{gcc/g++} version 7 or greater; \texttt{nvcc} version 11.
  \item \textbf{Data set:} Node classification benchmark data sets from OGB.
  \item \textbf{Run-time environment:} Ubuntu 18.04 (or modern linux distribution) with NVIDIA drivers installed.
  \item \textbf{Hardware:} NVIDIA GPU with sufficient memory. Distributed experiments require SLURM cluster with GPU nodes.
  \item \textbf{Experiments:} Single GPU performance comparisons, and distributed multi-GPU experiments
  \item \textbf{How much disk space required (approximately)?:} 100 GB for all experiments, 10 GB for a subset thereof.
  \item \textbf{How much time is needed to prepare workflow (approximately)?:} 1--2 hours with prior experience and access to hardware/clusters.
  \item \textbf{How much time is needed to complete experiments (approximately)?:} 1 hour for single GPU experiments and 4--12 hours for full set of distributed experiments.
  \item \textbf{Publicly available?:} Yes
  \item \textbf{Code licenses (if publicly available)?:} Apache License 2.0
  \item \textbf{Data licenses (if publicly available)?:} Amazon license and ODC-BY.
  \item \textbf{Archived (provide DOI)?:} \url{https://doi.org/10.5281/zenodo.6332979}
\end{itemize}

\subsection{Description}

\subsubsection{How delivered}

The code may be obtained from GitHub at \url{https://github.com/MITIBMxGraph/SALIENT_artifact}. Within the repository, scripts for streamlining the process of exercising the artifact are provided in the \texttt{experiments} directory. A dedicated readme file that documents the use of these scripts is provided at \url{https://github.com/MITIBMxGraph/SALIENT_artifact/blob/main/README.md}.

\subsubsection{Hardware dependencies}

The minimum requirements for exercising the software artifact are as follows.
The single GPU experiments require one NVIDIA GPU with sufficient memory, and one multi-core CPU that uses either the x86 or PowerPC architecture. We recommend using x86 CPUs as we have tested the installation process more thoroughly for them.

The distributed multi-GPU experiments require a SLURM cluster with GPU nodes. Such a cluster may be obtained through cloud services and accompanying software packages. For example, on Amazon Web Services one can use the ParallelCluster software to launch a SLURM cluster. 

Depending on the used hardware and available disk space, certain experiments may not be feasible. We have made an effort to reduce the disk space and memory requirements needed for running experiments on the largest data set, and we expect that GPUs with 16GB of memory and machines with 128GB of main memory will be able to run all or almost all of the experiments. For the distributed multi-GPU experiments, the PyG implementation often requires more than 128GB of memory when running on the ogbn-papers100M data set. Exercising the distributed experiments for PyG on this graph may require compute nodes with 256GB or more main memory. 

\subsubsection{Software dependencies}

Reasonably up-to-date NVIDIA drivers must be installed on the machine. For the distributed experiments, a SLURM cluster is required. The remaining software dependencies can be resolved using the \texttt{conda} package manager or by using the provided Dockerfile. If using docker, one must have \texttt{nvidia-docker} installed for GPU support within the container. 

\subsubsection{Data sets}

Graph data sets for node property prediction are taken from Open Graph Benchmark (OGB). To decrease the time and minimum hardware resources required for experiments, we have provided, as an option, the ability to download preprocessed versions of the graph data. If not electing to download the preprocessed graphs, the first execution of the code on a new graph will download it from OGB and perform preprocessing locally. 

\subsection{Installation}

We recommend referring to the installation instructions provided at \url{https://github.com/MITIBMxGraph/SALIENT_artifact/blob/main/README.md}. We summarize the installation process here.

\textbf{Installation using Docker:} We provide a docker container that can be used for running experiments on a single machine. Although the container could also be used to run distributed experiments, we have not tested this option. To use the docker container with NVIDIA GPUs, one should install docker and the NVIDIA Container Toolkit.

{\scriptsize
\begin{verbatim}
# Pull the container
docker pull nistath/salient:cuda-11.1.1

# Clone the code repository outside of the container
git clone \
    git@github.com:MITIBMxGraph/SALIENT_artifact.git

# Run docker container with host code folder mounted
docker run --ipc=host --gpus all -it \
    -v `pwd`/SALIENT_artifact:/salient \
    nistath/salient:cuda-11.1.1

# Install fast sampler
cd /salient/fast_sampler && python setup.py develop
\end{verbatim}
}

\textbf{Installation in Python environment:} We provide instructions to install the artifact in a Python environment. Such installation can be used for both the single GPU and distributed multi-GPU experiments (assuming access to a SLURM cluster). The instructions for installing in a conda environment are provided at \url{https://github.com/MITIBMxGraph/SALIENT_artifact/blob/main/INSTALL.md} and are summarized below.

{\scriptsize
\begin{verbatim}
# Install miniconda
wget https://repo.anaconda.com/miniconda/\
    Miniconda3-py38_4.10.3-Linux-x86_64.sh
bash Miniconda3-py38_4.10.3-Linux-x86_64.sh

# Create a conda environment for experiments
conda create -n salient python=3.8 -y
conda activate salient

# Install Pytorch, PyG, OGB, prettytable
conda install pytorch torchvision \
    torchaudio cudatoolkit=11.3 -c pytorch
conda install pyg -c pyg -c conda-forge
pip install ogb
conda install prettytable -c conda-forge

# Install patched PyTorch-Sparse
pip uninstall torch_sparse
FORCE_CUDA=1
pip install \
    git+git://github.com/rusty1s/pytorch_sparse.git@master

# Install fast_sampler
cd fast_sampler
python setup.py install
cd ..
\end{verbatim}
}

\subsection{Experiment workflow}

We recommend referring to the documentation for performing artifact evaluation located in the repository at \url{https://github.com/MITIBMxGraph/SALIENT_artifact/blob/main/README.md}. We summarize the experimental workflow here. Unless otherwise noted, all file paths are relative to the \texttt{experiments} directory in the repository. 

\textbf{Initial setup:} The script \texttt{initial\_setup.sh} can be executed to configure the number of sampling workers based on the hardware, and determine what data sets to download based on the available disk space. It will then, by default, download the appropriate preprocessed data sets.  

\textbf{Single GPU experiments:} We provide the script \texttt{run\_all\_single\_gpu\_experiments.sh} to run all single GPU experiments and display the final table of results. Additional scripts, documented in the artifact evaluation guide in the repository, are provided to run these experiments manually and regenerate the summary table of results.

\textbf{Distributed multi-GPU experiments:} These experiments require the use of a SLURM cluster. The file \texttt{all\_dist\_benchmarks.sh} must be modified to account for the configuration of the cluster at hand. Additional  instructions and guidance are provided in the artifact evaluation guide in the repository. The final table of results can be generated with the command:

{\scriptsize
\begin{verbatim}
python helper_scripts/parse_times.py \
    distributed_job_output/
\end{verbatim}
}

\subsection{Evaluation and expected result}

Upon completion of the single GPU experiments, a table will be produced that provides a performance breakdown of per-epoch runtime that reproduces the breakdowns provided in \tabref{bottleneck} and Figure~\ref{fig:perf.engineer} of the paper.

Upon completion of the distributed multi-GPU experiments, a table will be produced that provides the data needed to reproduce Figure~\ref{fig:GNNs}. Specifically, the table includes the per-epoch runtime and test accuracy for \oursystem and PyG across four GNN architectures shown in Figure~\ref{fig:GNNs}. The scripts may be modified to run on a different number of GPUs to reproduce the scalability experiment shown in Figure~\ref{fig:scaling}.

\subsection{Experiment customization}

The following experiment customizations are possible. The software may be directly exercised without the use of the dedicated artifact evaluation scripts. The scripts for single GPU experiments may be modified to use different GNN architectures, sampling fanouts, and hidden feature sizes by modifying the parameters in \texttt{performance\_breakdown\_config.cfg}.
The distributed multi-GPU experiments may be modified to run on different data sets and different numbers of GPUs.
The \texttt{fast\_sampler} extension can be integrated to other codes.

%
%
%

\section{Code Repository}
In addition to the artifact repository that focuses on benchmarking and reproducibility, an implementation of \oursystem for general-purpose usage is available at \url{https://github.com/MITIBMxGraph/SALIENT}.

\section{Architectures for Experiments}

See listings~\ref{lst:SAGE}, \ref{lst:GAT}, \ref{lst:GIN}, and \ref{lst:SAGERI} for the model definitions written in PyG.

\begin{figure*}[ht]
\begin{lstlisting}[basicstyle=\ttfamily\small,language=Python,label=lst:SAGE,caption=GraphSAGE model definition.]
def __init__(self, in_channels, hidden_channels, out_channels, num_layers):
  kwargs = dict(bias = False)
  conv_layer = SAGEConv
  super().__init__()
  self.num_layers = num_layers
  self.convs = torch.nn.ModuleList()
  self.hidden_channels = hidden_channels

  self.convs.append(conv_layer(in_channels, hidden_channels, **kwargs))
  for _ in range(num_layers - 2):
    self.convs.append(conv_layer(hidden_channels, hidden_channels, **kwargs))
  self.convs.append(conv_layer(hidden_channels, hidden_channels, **kwargs))
  self.reset_parameters()
  
def forward(self, x, adjs):
  end_size = adjs[-1][-1][1]
  for i, (edge_index, _, size) in enumerate(adjs):
    x_target = x[:size[1]]
    x = self.convs[i]((x, x_target), edge_index)
    if i != self.num_layers - 1:
      x = F.relu(x)
      x = F.dropout(x, p=0.5, training=self.training)
  return torch.log_softmax(x, dim=-1)
\end{lstlisting}
\end{figure*}

\begin{table*}[ht]
\begin{lstlisting}[basicstyle=\ttfamily\small,language=Python,label=lst:GAT,caption=GAT model definition.]
def __init__(self, in_channels, hidden_channels, out_channels, num_layers):
  kwargs = dict(bias = False, heads = 1)
  conv_layer = GATConv
  super().__init__()
  self.num_layers = num_layers
  self.convs = torch.nn.ModuleList()
  self.hidden_channels = hidden_channels
  
  self.convs.append(conv_layer(in_channels, hidden_channels, **kwargs))
  for _ in range(num_layers - 2):
    self.convs.append(conv_layer(hidden_channels, hidden_channels, **kwargs))
  self.convs.append(conv_layer(hidden_channels, out_channels, **kwargs))
  self.reset_parameters()
  
def forward(self, x, adjs):
  for i, (edge_index, _, size) in enumerate(adjs):
    x_target = x[:size[1]]
    x = self.convs[i]((x, x_target), edge_index)
    if i != self.num_layers - 1:
      x = F.relu(x)
      x = F.dropout(x, p=0.5, training=self.training)
  return torch.log_softmax(x, dim=-1)
\end{lstlisting}
\end{table*}

\begin{table*}[ht]
\begin{lstlisting}[basicstyle=\ttfamily\small,language=Python,label=lst:GIN,caption=GIN model definition.]
def __init__(self, in_channels, hidden_channels, out_channels, num_layers):
  kwargs = dict()
  conv_layer = GINConv
  super().__init__()
  self.num_layers = num_layers
  self.convs = torch.nn.ModuleList()
  self.hidden_channels = hidden_channels

  self.convs.append(GINConv(Sequential(Linear(in_channels, hidden_channels),
                    BatchNorm1d(hidden_channels), ReLU(),
                    Linear(hidden_channels, hidden_channels), ReLU())))
  for _ in range(num_layers - 2):
    self.convs.append(GINConv(Sequential(Linear(hidden_channels, hidden_channels),
                      BatchNorm1d(hidden_channels), ReLU(),
                      Linear(hidden_channels, hidden_channels), ReLU())))
  self.convs.append(GINConv(Sequential(Linear(hidden_channels, hidden_channels),
                    BatchNorm1d(hidden_channels), ReLU(), 
                    Linear(hidden_channels, hidden_channels), ReLU())))
  self.lin1 = Linear(hidden_channels, hidden_channels)
  self.lin2 = Linear(hidden_channels, out_channels)
  self.reset_parameters()
  
def forward(self, x, adjs):
  end_size = adjs[-1][-1][1]
  for i, (edge_index, _, size) in enumerate(adjs):
    x_target = x[:size[1]]
    x = self.convs[i]((x, x_target), edge_index)
  x = self.lin1(x).relu()
  x = F.dropout(x, p=0.5, training=self.training)
  x = self.lin2(x)
  return torch.log_softmax(x, dim=-1)
\end{lstlisting}
\end{table*}

\begin{table*}[ht]
\begin{lstlisting}[basicstyle=\ttfamily\small,language=Python,label=lst:SAGERI,caption=GragSAGE-RI model definition.]
def __init__(self, in_channels, hidden_channels, out_channels, num_layers):
  conv_layer = SAGEConv 
  kwargs = dict(bias = False)
  super().__init__()
  self.num_layers = num_layers
  self.convs = torch.nn.ModuleList()
  self.bns = torch.nn.ModuleList()
  self.res_linears = torch.nn.ModuleList()
  self.hidden_channels = hidden_channels

  self.convs.append(conv_layer(in_channels, hidden_channels, **kwargs))
  self.bns.append(torch.nn.BatchNorm1d(hidden_channels))
  self.res_linears.append(torch.nn.Linear(in_channels, hidden_channels))
  for _ in range(num_layers - 2):
    self.convs.append(conv_layer(hidden_channels, hidden_channels, **kwargs))
    self.bns.append(torch.nn.BatchNorm1d(hidden_channels))
    self.res_linears.append(torch.nn.Identity())
  self.convs.append(conv_layer(hidden_channels, hidden_channels, **kwargs))
  self.bns.append(torch.nn.BatchNorm1d(hidden_channels))
  self.res_linears.append(torch.nn.Identity())
  
def forward(self, _x, adjs):
  collect = []
  end_size = adjs[-1][-1][1]
  x = F.dropout(_x, p=0.1, training=self.training)
  collect.append(x[:end_size])
  for i, (edge_index, _, size) in enumerate(adjs):
    x_target = x[:size[1]]
    x = self.convs[i]((F.dropout(x,p=0.1,training=self.training),
                       F.dropout(x_target, p=0.1, training=self.training)), edge_index)
    x = self.bns[i](x)
    x = F.leaky_relu(x)
    x = F.dropout(x, p=0.1, training=self.training)
    collect.append(x[:end_size])
    x += self.res_linears[i](x_target)
  return torch.log_softmax(self.mlp(torch.cat(collect, -1)), dim=-1)
\end{lstlisting}
\end{table*}

\end{document}